\documentclass{article} 
\usepackage{iclr2027_conference,times}

\usepackage{amsmath}
\usepackage{amssymb}
\usepackage{mathtools}
\usepackage{mathrsfs}

\usepackage{graphicx}
\usepackage{subcaption}
\usepackage{booktabs}
\usepackage{microtype}
\usepackage{enumitem}

\usepackage{algorithm}
\usepackage{algpseudocode}
\usepackage{booktabs}
\usepackage{multirow}

\usepackage{tikz}
\usetikzlibrary{decorations.pathreplacing}

\IfFileExists{bbm.sty}{\usepackage{bbm}}{}
\providecommand{\mathbbm}[1]{\mathbf{#1}}

\DeclareMathOperator*{\argmin}{arg\,min}

\newtheorem{theorem}{Theorem}
\newtheorem{proposition}{Proposition}
\newtheorem{lemma}{Lemma}
\newtheorem{corollary}{Corollary}

\newtheorem{assumption}{Assumption}

\newenvironment{proof}{\par\medskip\noindent\textit{Proof.}\ }{\hfill$\square$\par\medskip}

\newcommand{\blfootnote}[1]{%
  \begingroup
  \renewcommand{\thefootnote}{}%
  \footnotetext{#1}%
  \addtocounter{footnote}{-1}%
  \endgroup
}

\usepackage{hyperref}
\usepackage{url}

\title{Hierarchical Distance Decomposition for\\Minimum-Action-Distance Estimation}

\author{
Lorenzo Steccanella$^{}$\thanks{Corresponding author: \texttt{lorenzo.steccanella.w@gmail.com}},
Joshua~B.~Evans$^2$,
Amir Esterhuysen$^1$,
Anders Jonsson$^1$\\[0.5em]
$^1$Department of Information and Communication Technologies,\\
Universitat Pompeu Fabra, Barcelona, Spain\\[0.3em]
$^2$Department of Computer Science,\\ 
University of Bath, Bath, United Kingdom
}

\iclrfinalcopy 

\begin{document}
\maketitle
\blfootnote{Preprint.}

\fancyhead{}
\renewcommand{\headrulewidth}{0pt}
\cfoot{\thepage}

\begin{abstract}
Estimating how many actions are needed for an agent to move between distant states is difficult. The Minimum Action Distance (MAD) formalizes this quantity, but approximations learned from experience can be locally accurate yet unreliable over long horizons. We introduce \emph{hierarchical distance decomposition}, a post-hoc method that improves long-range estimates by composing local predictions through a sparse graph of landmark states. It can be applied to any existing learned distance model without retraining. We present theoretical results that bound errors in the resulting distance estimates and establish order preservation for sufficiently separated state pairs. Empirically, the method improves both long-range distance estimation accuracy and downstream planning performance.
\end{abstract}

\section{Introduction}
\label{sec:intro}

For an autonomous agent, knowing the distance between two states can support many aspects of learning and decision-making, including exploration, goal-directed control, planning, reward shaping, and subgoal selection. One particularly direct distance measure is the \emph{Minimum Action Distance} (MAD), defined as the minimum number of actions required for an agent to transition from one state to another \citep{steccanella2026}. In deterministic environments, the MAD is the shortest-path distance in the state transition graph. In stochastic environments, it is the shortest-path distance in the support of the transition function, and acts as a lower bound on the expected number of actions required to reach a target state.

In practice, computing the MAD directly is generally impractical, so it must be learned from experience. Existing approaches learn a distance function, often defined as a quasimetric, over a learned state representation. A key limitation of existing approaches is that such approximations can produce accurate distance estimates for nearby state pairs but become unreliable over long horizons, primarily due to error accumulation, limited representation capacity, and distributional shifts.

We propose \emph{hierarchical distance decomposition}, a method for estimating long-range distances by composing short-range predictions from an existing learned distance model. The method selects a sparse set of \emph{landmark states} from a dataset of experience and builds a directed graph whose edges connect pairs of landmarks that the model estimates to be close. For a state pair $(s,g)$, it connects $s$ and $g$ to nearby landmarks and estimates their distance using a shortest path through the landmark graph. In this way, the method replaces a potentially unreliable long-range prediction with a sequence of more reliable local predictions. The method is agnostic to the choice of base learned distance model, is applied post hoc, and requires no retraining of the base model or underlying representation.

Our contributions are: (1) a hierarchical estimator that uses only short-range evaluations of an existing learned distance model to answer long-range queries (Section~\ref{sec:algorithm}); (2) theoretical results that conditionally bound under- and over-estimation and establish an order-preservation result for sufficiently separated state pairs (Section~\ref{sec:theory}); and (3) empirical results showing improvements in both long-range distance estimation and downstream planning performance when applied across multiple base distance models, environments, and experience datasets (Section~\ref{sec:results}).


\section{Related Work}
\label{sec:related}


\paragraph{Temporal distance learning.}
Many reinforcement learning methods use notions of temporal distance that reflect how long it may take an agent to transition from one state to another. Early work explored the connection between goal-conditioned value functions and the cost of reaching different goals~\citep{kaelbling1993}, while later work defined temporal distances more explicitly. \citet{hartikainen2020} learn a policy-dependent temporal distance based on the \emph{expected} number of decision stages required to reach one state from another, while \citet{steccanella2022} alternatively define the Minimum Action Distance (MAD) as the \emph{minimum} number of actions required to do so.

Several earlier methods embed states into a learned latent space where Euclidean distance captures temporal distance~\citep{steccanella2022,park2024metra,park2024foundation}. However, Euclidean distances are symmetric, whereas temporal distances such as the MAD may not be. To capture this asymmetry, more recent work has used quasimetrics to represent temporal distance functions. \citet{wang2023optimal} exploit the quasimetric structure of optimal goal-conditioned value functions, while \citet{steccanella2026} use quasimetric representations to directly approximate the MAD from offline state trajectories. Related quasimetric methods recover optimal goal-conditioned policies from suboptimal and stochastic offline data~\citep{myers2025} and use multi-step updates to better combine local and global distance information~\citep{zheng2026}.

\paragraph{Graph-based planning.}
Graph search has been used to compose local information over longer horizons, often to support long-horizon control. \citet{eysenbach2019} construct a graph over observed states using distances derived from a goal-conditioned value function, and use shortest paths through this graph to select intermediate subgoals. Later work extends this idea in several ways: \citet{baek2025} cluster states from offline trajectories using a learned representation that captures temporal distance, and search the resulting graph to select subgoals; \citet{opryshko2026testtime} bias graph search towards paths composed of more reliable local transitions, retaining longer-distance edges only when necessary; and \citet{kobanda2026} construct a landmark graph using a learned distance function and update it online when observed traversal times disagree with its predictions.

The closest prior methods to ours use graphs to directly improve learned distance or value estimates. \citet{yang2020} use shortest paths over a graph built using learned local distances to generate long-horizon distance targets, and distill these into a global distance representation. Similarly, \citet{bagatella2023} aggregate shorter-horizon estimates along shortest paths through a graph to replace unreliable long-horizon estimates for model-based planning.

\paragraph{Hierarchical reinforcement learning.}
A related line of work uses hierarchical reinforcement learning to decompose long-horizon tasks, often using learned distances or values to structure their hierarchies. These methods variously learn higher- and lower-level control policies~\citep{park2023,canesse2025,wibault2026}, temporally abstract value functions~\citep{ahn2026}, or abstract models of the environment~\citep{rodriguezsanchez2024,piriyakulkij2026}.

\paragraph{Relation to our work.}
Hierarchical distance decomposition differs from prior work in two main ways. First, we use graph-based decomposition directly to improve learned distance estimates, rather than to support any particular downstream use. Second, we apply it post-hoc to an existing learned distance model, without retraining the model or its underlying representation. Temporal-distance methods focus on how distance models themselves are learned, while hierarchical methods focus on how to introduce hierarchy to improve policies, value functions, or models to support planning and control. The closest graph-based approaches also differ on these points: \citet{yang2020} retrain a global representation using graph-derived targets, while \citet{bagatella2023} correct goal-conditioned value estimates specifically for planning.

\section{Problem Setting}
\label{sec:problem}

We consider a Markov Decision Process (MDP) $\mathcal{M}=\langle\mathcal{S},\mathcal{A},\mathrm{P},\mathrm{R},\gamma\rangle$, where $\mathcal{S}$ is a set of states, $\mathcal{A}$ is a set of actions, $\mathrm{P}:\mathcal{S}\times\mathcal{A}\times\mathcal{S}\rightarrow\left[0,1\right]$ is a transition function, $\mathrm{R}:\mathcal{S}\times\mathcal{A}\times\mathcal{S}\rightarrow\mathbb{R}$ is a reward function, and $\gamma\in\left[0,1\right]$ is a discount factor. At each decision stage, an agent observes a state $s\in\mathcal{S}$, selects an action according to a policy $\pi:\mathcal{S}\times\mathcal{A}\rightarrow\left[0,1\right]$, and subsequently transitions to a next-state $s'\in\mathcal{S}$ according to $\mathrm{P}$ and receives a reward $r\in\mathbb{R}$ according to $\mathrm{R}$. We assume access to a dataset $\mathcal{D}$ of state-only trajectories collected by an agent following an unknown behavior policy.

Let $d^*:\mathcal{S}\times\mathcal{S}\rightarrow\mathbb{R}_{\geq0}$ denote the target distance we wish to estimate. Although our construction applies more generally, we consider $d^*=d_{\mathrm{MAD}}$ in this paper, where the MAD is the minimum number of actions required to move from one state to another \citep{steccanella2026}. We assume that $d^*$ is a quasimetric: it is non-negative and satisfies the triangle inequality for all state pairs, but need not be symmetric.
We also assume access to a learned function $d_\theta : \mathcal{S} \times \mathcal{S} \to \mathbb{R}_{\geq 0}$ that approximates $d^*$. We make no assumptions about how $d_\theta$ was trained. Our goal is to use the dataset of offline trajectories $\mathcal{D}$ and existing distance model $d_\theta$ to obtain more accurate long-range distances than using $d_\theta$ alone, without requiring any retraining of $d_\theta$.

Learned distance models may be reliable locally, but become less accurate over longer horizons~\citep{wang2023optimal, park2024foundation, steccanella2026}. Three factors contribute to this degradation. First, \emph{errors may accumulate}  as distance information is propagated across many local relations during training. Second, \emph{finite-capacity models} may compress a wide range of high-magnitude distance values into similar outputs. Third, offline data collected through local exploration often \emph{differs in distribution} from the distant state pairs whose distances may be queried at test time.
Hierarchical distance decomposition addresses these problems by composing a model's more accurate short-range estimates to obtain improved long-range estimates.


\section{Hierarchical Distance Decomposition}
\label{sec:algorithm}
We present pseudocode for hierarchical distance decomposition in Algorithm~\ref{alg:hierarchical}. The method consists of two phases: \emph{construction} (\textsc{BuildHierarchy}) and \emph{querying} (\textsc{QueryDistance}). During construction, the method selects a set of landmark states from a dataset, connects nearby landmarks in a directed graph, and precomputes all-pairs shortest-path distances. At query time, these precomputed distances are combined with local predictions to estimate distances between arbitrary state pairs.

\begin{algorithm}[t]
\small
\caption{Hierarchical Distance Decomposition}
\label{alg:hierarchical}
\begin{algorithmic}[1]

\Procedure{BuildHierarchy}{$\mathcal{D}, d_\theta, \delta, \kappa$}
    \State \textbf{Input:} Dataset of state trajectories $\mathcal{D} \subset \mathcal{S}$; base quasimetric $d_\theta$; spacing parameter $\delta > 0$; edge-threshold multiplier $\kappa > 1$
    \State \textbf{Output:} Landmark set $\mathcal{C}$; all-pairs shortest-path matrix $D_h$
    \State $\mathcal{C} \gets \emptyset$
    \For{each state $s$ in $\mathcal{D}$} \Comment{Greedy two-sided $\delta$-net over the dataset}
        \If{$\mathcal{C} = \emptyset$ \textbf{ or } $\min_{c \in \mathcal{C}} \max\!\bigl(d_\theta(s, c),\; d_\theta(c, s)\bigr) > \delta$}
            \State $\mathcal{C} \gets \mathcal{C} \cup \{s\}$ \Comment{Retire $s$ only if one landmark is close \emph{both} ways}
        \EndIf
    \EndFor
    \State Construct directed graph $G_\mathcal{C} = (\mathcal{C}, E_\mathcal{C})$ with $E_\mathcal{C} = \bigl\{(c_i, c_j) : d_\theta(c_i, c_j) \le \kappa\delta\bigr\}$ and $w(c_i, c_j) = d_\theta(c_i, c_j)$ \Comment{Keep only short edges: long-range predictions are never trusted}
    \State $D_h \gets \textsc{FloydWarshall}(G_\mathcal{C})$ \Comment{All-pairs shortest paths}
    \State \Return $\mathcal{C}, D_h$
\EndProcedure

\vspace{0.5em}

\Procedure{QueryDistance}{$s_0, s_g, d_\theta, \mathcal{C}, D_h, \kappa, k$}
    \State \textbf{Input:} Query pair $(s_0, s_g)$; base quasimetric $d_\theta$; landmark set $\mathcal{C}$; shortest-path matrix $D_h$; edge-threshold multiplier $\kappa$; neighbourhood size $k$
    \State \textbf{Output:} Distance estimate $\hat{d}_H(s_0, s_g)$
    \State $c_s^* \gets \argmin_{c \in \mathcal{C}}\, d_\theta(s_0, c)$ \Comment{Nearest forward landmark}
    \State $c_g^* \gets \argmin_{c \in \mathcal{C}}\, d_\theta(c, s_g)$ \Comment{Nearest backward landmark}
    \If{$c_s^* = c_g^*$} \Comment{Same-landmark regime: too close to route}
        \State \Return $d_\theta(s_0, s_g)$
    \EndIf
    \State $\mathcal{C}_s \gets \{c \in \mathcal{C} : d_\theta(s_0, c) \le \kappa\delta\}$ \Comment{Nonempty by Proposition~\ref{prop:covering}; truncated to the $k$ nearest}
    \State $\mathcal{C}_g \gets \{c \in \mathcal{C} : d_\theta(c, s_g) \le \kappa\delta\}$ \Comment{Nonempty by Proposition~\ref{prop:covering}; truncated to the $k$ nearest}
    \State \Return $\displaystyle\min_{c_s \in \mathcal{C}_s,\; c_g \in \mathcal{C}_g}\; d_\theta(s_0, c_s) + D_h(c_s, c_g) + d_\theta(c_g, s_g)$
\EndProcedure

\end{algorithmic}
\end{algorithm}

\subsection{Constructing the Landmark Graph}
\label{sec:construction}

The algorithm begins by using the state-only trajectory dataset $\mathcal{D}$ and an existing learned distance function $d_\theta$ to generate a set of landmarks $\mathcal{C}$. It processes the states in $\mathcal{D}$ greedily, adding a state $s$ as a new landmark only when no existing landmark $c\in\mathcal{C}$ satisfies both $d_\theta(s,c)\leq\delta$ and $d_\theta(c,s)\leq\delta$. The parameter $\delta$ therefore controls the spacing between landmarks. This two-sided requirement is important because $d_\theta$ may be asymmetric: a landmark that is close when moving from $s$ to $c$ need not be close when moving from $c$ to $s$. We say that a landmark $c$ \emph{covers} a state $s$ when both conditions hold. This construction ensures the following covering property.

\begin{proposition}[Covering property]
\label{prop:covering}
After \textsc{BuildHierarchy} terminates, every state $s \in \mathcal{D}$ has a landmark $c \in \mathcal{C}$ with
\[
    \max\bigl(d_\theta(s, c),\; d_\theta(c, s)\bigr) \;\le\; \delta .
\]
\end{proposition}
\begin{proof}
Algorithm~\ref{alg:hierarchical} visits every state in $\mathcal{D}$. Any state not covered by the landmarks selected so far is added to $\mathcal{C}$ and therefore covers itself with zero distance in both directions. Since landmarks are never removed, this coverage is preserved.
\end{proof}

\paragraph{Why coverage is two-sided.}
To answer a distance query for a starting state $s_0$ and a goal state $s_g$, the start state must connect to a landmark $c_0$ through $d_\theta(s_0,c_0)$, while a landmark $c_g$ must connect to the goal through $d_\theta(c_g,s_g)$. Since any state may appear as either the start or goal, the covering condition requires each state to have a landmark that can support both roles.


Consider a cheaper alternative (which we do not use) that constructs two separate one-sided landmark sets: one ensures coverage from each state to a landmark, the other ensures coverage from a landmark to each state, and the method takes their union. However, different landmarks may satisfy these two conditions. The resulting set therefore does not guarantee the directed connectivity required to answer a query; Appendix~\ref{app:max-rule} gives an example in which this alternative produces an infinite distance estimate even when $d_\theta=d^*$. Our two-sided rule avoids this failure by requiring a single landmark to support both roles, at the cost of selecting more landmarks.


\paragraph{The landmark graph.}
After obtaining the landmark set $\mathcal{C}$, we construct a directed graph over its elements. We retain a directed edge $(c_i,c_j)$ only when $d_\theta(c_i,c_j)\le\kappa\delta$, assigning it weight $d_\theta(c_i,c_j)$. Applying Floyd--Warshall to this graph gives the matrix $D_h$, where $D_h(c_i,c_j)$ is the length of the shortest path from $c_i$ to $c_j$, or infinity if no such path exists. We compute $D_h$ once offline and reuse it for each query.
On a fully connected graph, shortest-path search could select a single long edge that the base model severely underestimates, recreating the problem we aimed to avoid. By retaining only edges with predicted distance at most $\kappa\delta$, every finite distance in $D_h$ is a sum of short-range predictions. The multiplier $\kappa$ controls the resulting trade-off: lower values favour more local predictions, while higher values raise connectivity at the cost of retaining longer-range edges.

\subsection{Querying Distances}
\label{sec:query}

For a query $(s_0,s_g)$, \textsc{QueryDistance} considers routes of the form $s_0\to c_s\to\cdots\to c_g\to s_g$.
We distinguish \emph{edges}, which connect pairs of landmarks and have stored weights $d_\theta(c_i, c_j)$ in $D_h$, from \emph{legs}, which connect a query state to a landmark and whose distances are evaluated at query time. Each edge or leg is restricted to model distances of at most $\kappa\delta$, so the overall estimate consists only of short-range predictions from the base model $d_\theta$.
A candidate route therefore consists of an entry leg from $s_0$ to $c_s$, a shortest landmark path from $c_s$ to $c_g$ with cost $D_h(c_s, c_g)$, and an exit leg from $c_g$ to $s_g$. We denote the closest landmarks to the state and goal in the relevant directions by 
\[
    c_s^* = \argmin_{c \in \mathcal{C}} d_\theta(s_0, c),
    \qquad
    c_g^* = \argmin_{c \in \mathcal{C}} d_\theta(c, s_g)
\]
for the landmark closest to the start in the forward direction and the landmark closest to the goal in the backward direction. The resulting hierarchical distance estimate is
\begin{equation}
\label{eq:hier_estimate}
    \hat{d}_H(s_0, s_g) =
    \begin{cases}
        d_\theta(s_0, s_g)
        & \text{if } c_s^* = c_g^*, \\[6pt]
        \displaystyle\min_{c_s \in \mathcal{C}_s,\, c_g \in \mathcal{C}_g}\;
        d_\theta(s_0, c_s) + D_h(c_s, c_g) + d_\theta(c_g, s_g)
        & \text{otherwise}.
    \end{cases}
\end{equation}

For each query, we consider sets of candidate entry and exit landmarks. Entry candidates are landmarks within $\kappa\delta$ of $s_0$ in the forward direction, while exit candidates are landmarks within $\kappa\delta$ of $s_g$ in the backward direction. In practice, we truncate each set to its $k$ nearest landmarks, with $k=5$. For states in $\mathcal{D}$, Proposition~\ref{prop:covering} ensures that both sets are nonempty by supplying a landmark within $\delta$ in both directions. Considering multiple candidates is important because the nearest entry and exit landmarks do not necessarily yield the cheapest complete route.

The first case of Equation~\ref{eq:hier_estimate} handles state pairs that share the same closest landmark. For such pairs, routing through the graph can introduce an $O(\delta)$ detour that dominates their true separation. So, the estimator simply returns the direct local prediction $d_\theta(s_0,s_g)$ instead.

\subsection{Computational Cost}
\label{sec:comp_cost}

In this section, we highlight the computational cost of performing the proposed Hierarchical Distance Decomposition. We distinguish between the costs of \emph{construction} (\textsc{BuildHierarchy}) and \emph{querying} (\textsc{QueryDistance}). Let $|\mathcal{D}|$ denote the number of states in the dataset and $|\mathcal{C}|$ the number of landmarks selected from it.

The \emph{construction} phase is where most of the computational overhead occurs. Selecting landmarks parses the dataset of trajectories once and tests each state against the landmarks retained so far, for at most $O(|\mathcal{D}|\; |\mathcal{C}|)$ evaluations of $d_\theta$. Building the graph evaluates $d_\theta$ on each ordered pair of landmarks to find the edges below $\kappa\delta$, at a cost of $O(|\mathcal{C}|^2)$. Floyd--Warshall then turns the pruned graph into $D_h$ using $O(|\mathcal{C}|^3)$ operations.

The \emph{querying} phase needs $d_\theta(s_0,c)$ and $d_\theta(c,s_g)$ for every landmark, which is $2|\mathcal{C}|$ evaluations, and those values determine $c_s^*$, $c_g^*$, and both candidate sets. The three base models make those evaluations much cheaper than the count suggests. Each computes $d_\theta$ from a learned representation $\phi$ (Appendix~\ref{app:base-models}), as either $d_{simple}$ or $\|\cdot\|_2$, so we store $\phi(c)$ for every landmark during construction. A query then encodes only $s_0$ and $s_g$ and evaluates the distance head against that stored matrix in a single batched operation.

\section{Theoretical Analysis}
\label{sec:theory}

Under two assumptions---local accuracy of the base model and coverage of relevant paths by the dataset---we recover a two-sided bound on estimator error:
\[
    \mathrm{err}(s_0,s_g) \;=\; \hat d_H(s_0,s_g) \;-\; d^*(s_0,s_g)
\]
Theorem~\ref{thm:approx} shows that
\[
    d^*(s_0,s_g) - (m+2)\,\varepsilon
    \;\le\; \hat d_H(s_0,s_g) \;\le\;
    d^*(s_0,s_g) + M\,(2\delta + 3\varepsilon) + 2\delta ,
\]
where $\varepsilon$ bounds the local error of the base model, $m$ is the number of graph edges on the route the estimator returns, and $M=\lceil d^*(s_0,s_g)/\delta\rceil$ is the number of hops of the route used for comparison. The left inequality says the estimate can fall below the target distance only by the local error accumulated along the returned route. The right inequality charges a fixed cost per hop of the route used for comparison: $2\delta$ of detour and $3\varepsilon$ of conversion, plus $2\delta$ for the two legs. Throughout, $d^*$ is a quasimetric: $d^*(x,x)=0$, $d^*\ge0$, and $d^*(x,z)\le d^*(x,y)+d^*(y,z)$. We make no triangle-inequality assumption about $d_\theta$.

\paragraph{Units.}
We express $d^*$ in the same units as $d_\theta$, allowing $d^*$, $d_\theta$, $\delta$, and $\varepsilon$ to use a common scale. This is without loss of generality: rescaling the model by a positive constant rescales both sides of every condition below. In particular, $\delta$ is a threshold on the model output. Applying the same numerical $\delta$ to models with different output scales produces landmark graphs of different resolution.

\subsection{Assumptions}
\label{sec:assumptions}

The analysis rests on two assumptions: the model must be accurate on every evaluated local pair (Assumption~\ref{ass:local}), and landmarks must cover the path used for comparison (Assumption~\ref{ass:coverage}). Only the overestimation side uses the second one, as the proof idea below explains.

\begin{assumption}[Local accuracy]
\label{ass:local}
There is an $\varepsilon \in [0, \delta]$ such that, for all $(x, y) \in \mathcal{S}\times\mathcal{S}$,
\[
    \min\bigl(d_\theta(x, y),\; d^*(x, y)\bigr) \;\le\; \kappa\delta
    \quad\Longrightarrow\quad
    \bigl|d_\theta(x, y) - d^*(x, y)\bigr| \;\le\; \varepsilon .
\]
\end{assumption}

This ensures that when either the model or target distance classifies a pair as local, their values differ by at most $\varepsilon$. Because the condition uses a minimum, the assumption applies in the following ways:
\begin{subequations}
\label{eq:two-readings}
\begin{align}
    d_\theta(x, y) \le \kappa\delta
    &\;\Longrightarrow\;
    d^*(x, y) \;\le\; d_\theta(x, y) + \varepsilon ,
    \label{eq:no-shortcut}
    \\
    d^*(x, y) \le \kappa\delta
    &\;\Longrightarrow\;
    d_\theta(x, y) \;\le\; d^*(x, y) + \varepsilon .
    \label{eq:survives}
\end{align}
\end{subequations}
Equation~\ref{eq:no-shortcut} excludes \emph{false shortcuts}: a retained edge classified as short by the model is also short in the target distance, up to $\varepsilon$. Equation~\ref{eq:survives} ensures that a genuinely short pair is not removed by pruning. The first property is particularly important, as shortest-path search actively selects underestimated edges. Pairs large under both $d_\theta$ and $d^*$ are unconstrained; their predictions are never used in an estimate.

Although the assumption is stated for all $(x,y)\in\mathcal{S}\times\mathcal{S}$, it is only applied to the following pair types: two landmarks, a query state and a landmark, a state of the reference path with its covering landmark. Landmarks come from $\mathcal{D}$, so the only pairs that reach beyond the dataset are the two query legs and the snaps along the reference path. Accuracy elsewhere in $\mathcal{S}$ is irrelevant to the bounds.

The second assumption concerns data coverage rather than model accuracy. It requires landmarks near the reference paths, but says nothing about the rest of $\mathcal{S}$.

\begin{assumption}[Coverage]
\label{ass:coverage}
For every query pair $(s_0, s_g)$, some shortest $d^*$-path from $s_0$ to $s_g$ stays within $\delta$ of $\mathcal{C}$: every state $s$ on this path has a landmark $c \in \mathcal{C}$ with
\[
    \max\bigl(d_\theta(s, c),\; d_\theta(c, s)\bigr) \;\le\; \delta .
\]
\end{assumption}

Proposition~\ref{prop:covering} gives this condition only for states in $\mathcal{D}$, so it discharges the assumption solely on the part of the path that the dataset contains. It says nothing about unseen states: for $s_0,s_g\notin\mathcal{D}$ nothing prevents $\min_{c\in\mathcal{C}} d_\theta(s_0,c)>\delta$ or $\min_{c\in\mathcal{C}} d_\theta(c,s_g)>\delta$. Coverage of the two query states is therefore assumed here rather than inherited from the construction. In practice, the assumption ensures some shortest path through covered states can be traced for each query, and is only required along such paths.

Every evaluation of $d_\theta$ entering the estimate, edge weights and legs alike, is on a pair at model distance at most $\kappa\delta$, so Assumption~\ref{ass:local} applies to all of them (Lemma~\ref{lem:all-local} in Appendix~\ref{app:proofs}).

\paragraph{Proof idea.}
The two inequalities are proved in opposite ways.

\emph{No underestimation.} The route the estimator returns is itself a chain $s_0\to c_0\to\cdots\to c_m\to s_g$, and $\hat d_H(s_0,s_g)$ is the sum of its $m+2$ terms. Every term is a local evaluation of $d_\theta$ (Lemma~\ref{lem:all-local}), so Assumption~\ref{ass:local} places it within $\varepsilon$ of the target distance of the corresponding segment. Those segments form a path from $s_0$ to $s_g$, so the triangle inequality for $d^*$ bounds the sum of their target lengths from below by $d^*(s_0,s_g)$, which gives the left inequality. Coverage plays no part here, since the argument uses only the route that is actually returned.

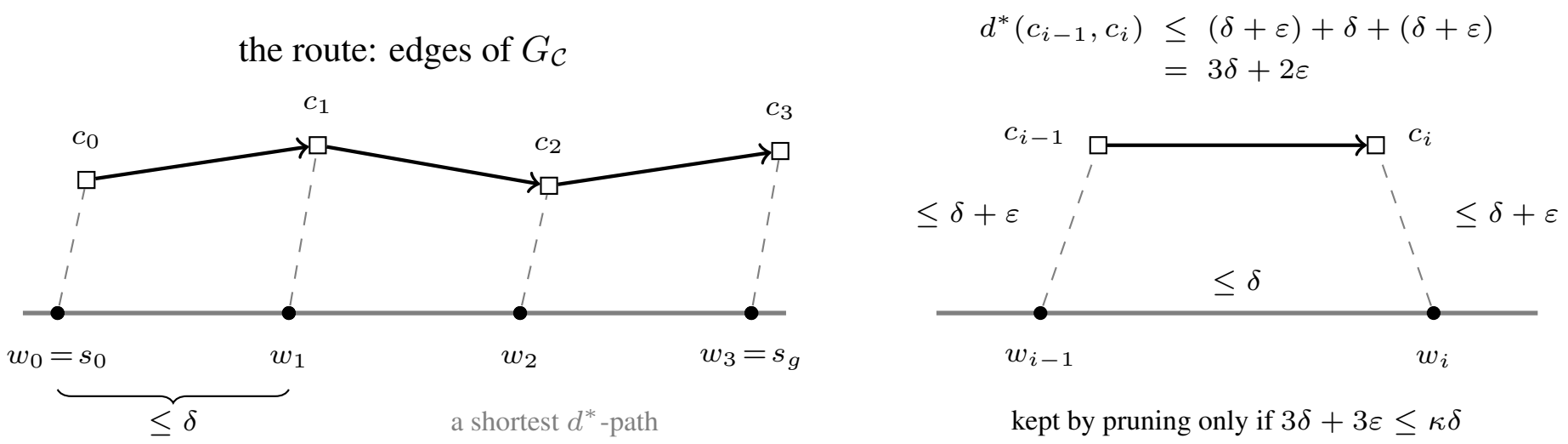
\begin{figure}[t]
\centering
\begin{tikzpicture}[
  scale=0.90,
  transform shape,
  wp/.style={circle, fill=black, inner sep=1.2pt},
  lm/.style={rectangle, draw, fill=white, inner sep=2pt},
  snap/.style={dashed, gray},
  lbl/.style={font=\small},
  txt/.style={font=\scriptsize},
  brc/.style={decorate, decoration={brace, amplitude=3pt}}
]
\begin{scope}
\draw[gray, line width=1pt] (-0.3,0) -- (6.3,0);
\foreach \x in {0, 2.0, 4.0, 6.0} \node[wp] at (\x,0) {};
\node[wp] (x0) at (0,0) {};
\node[wp] (x1) at (2.0,0) {};
\node[wp] (x2) at (4.0,0) {};
\node[wp] (x3) at (6.0,0) {};
\node[txt] at (0,-0.40) {$w_0\!=\!s_0$};
\node[txt] at (2.0,-0.40) {$w_1$};
\node[txt] at (4.0,-0.40) {$w_2$};
\node[txt] at (6.0,-0.40) {$w_3\!=\!s_g$};
\node[lm] (c0) at (0.25,1.15) {};
\node[lm] (c1) at (2.25,1.45) {};
\node[lm] (c2) at (4.25,1.10) {};
\node[lm] (c3) at (6.25,1.40) {};
\node[txt] at (0.25,1.50) {$c_0$};
\node[txt] at (2.25,1.80) {$c_1$};
\node[txt] at (4.25,1.45) {$c_2$};
\node[txt] at (6.25,1.75) {$c_3$};
\draw[snap] (x0) -- (c0);
\draw[snap] (x1) -- (c1);
\draw[snap] (x2) -- (c2);
\draw[snap] (x3) -- (c3);
\draw[->, thick] (c0) -- (c1);
\draw[->, thick] (c1) -- (c2);
\draw[->, thick] (c2) -- (c3);
\draw[brc] (2.0,-0.66) -- (0,-0.66);
\node[txt] at (1.0,-0.95) {$\le \delta$};
\node[txt, gray] at (4.3,-0.95) {a shortest $d^*$-path};
\node[lbl] at (3.0,2.25) {the route: edges of $G_\mathcal{C}$};
\end{scope}
\begin{scope}[xshift=7.3cm]
\draw[gray, line width=1pt] (0.3,0) -- (5.5,0);
\node[wp] (y0) at (1.2,0) {};
\node[wp] (y1) at (4.6,0) {};
\node[txt] at (1.2,-0.40) {$w_{i-1}$};
\node[txt] at (4.6,-0.40) {$w_i$};
\node[lm] (d0) at (1.7,1.45) {};
\node[lm] (d1) at (4.1,1.45) {};
\node[txt, anchor=east] at (1.55,1.52) {$c_{i-1}$};
\node[txt, anchor=west] at (4.25,1.52) {$c_i$};
\draw[snap] (y0) -- (d0);
\draw[snap] (y1) -- (d1);
\node[txt, anchor=east] at (1.15,0.85) {$\le \delta+\varepsilon$};
\node[txt, anchor=west] at (4.65,0.85) {$\le \delta+\varepsilon$};
\node[txt] at (2.9,0.26) {$\le \delta$};
\draw[->, thick] (d0) -- (d1);
\node[txt] at (2.9,2.45) {$d^*(c_{i-1},c_i) \;\le\; (\delta+\varepsilon) + \delta + (\delta+\varepsilon)$};
\node[txt] at (2.9,2.10) {$=\; 3\delta + 2\varepsilon$};
\node[txt] at (2.9,-0.95) {kept by pruning only if $3\delta + 3\varepsilon \le \kappa\delta$};
\end{scope}
\end{tikzpicture}
\caption{The route exhibited for the overestimation bound.  \emph{Left:} waypoints (dots) sit on a true shortest path with spacing $\delta$. Each is snapped to a landmark (squares) beside it through which the route runs, never through waypoints.  \emph{Right:} one hop, the source of each $\varepsilon$---contained only by the two snaps. Coverage is stated in $d_\theta$, so converting a snap into $d^*$ costs $\varepsilon$ apiece.  The gap carries none, as waypoints are spaced with $d^*$ by default.  This gives $3\delta + 2\varepsilon$ instead of $3\delta + 3\varepsilon$.  The third $\varepsilon$ only appears later, converting the finished edge back into the model's units, the price $3\delta+3\varepsilon$ that pruning tests against $\kappa\delta$.}
\label{fig:construction}
\end{figure}

\emph{Bounded overestimation.} In this direction the returned route is unknown, so we use that the estimate is a minimum over feasible routes: it is enough to exhibit one route and bound its cost, because the estimator's own answer cannot be larger. Assumption~\ref{ass:coverage} supplies such a route. Along a shortest $d^*$-path we mark waypoints $w_0,\dots,w_M$ with $w_0=s_0$ and $w_M=s_g$ at spacing at most $\delta$, and replace each waypoint $w_i$ by a landmark $c_i$ that covers it, so that the route runs through the landmarks rather than through the waypoints (Figure~\ref{fig:construction}, left). Each of its edges is then priced through the two waypoints it spans (Figure~\ref{fig:construction}, right),
\begin{equation}
\label{eq:snap}
    d^*(c_{i-1},c_i) \;\le\; \underbrace{d^*(c_{i-1},w_{i-1})}_{\text{snap}} + \underbrace{d^*(w_{i-1},w_i)}_{\text{progress}} + \underbrace{d^*(w_i,c_i)}_{\text{snap}} ,
\end{equation}
one step of progress along the reference path plus two snaps of size at most $\delta$. Summed over the $M$ edges, the progress terms telescope to $d^*(s_0,s_g)$, while each hop is charged $2\delta$ for its two snaps, and the two legs add a further $2\delta$: this is the per-hop cost in the right inequality. The right panel of Figure~\ref{fig:construction} shows where each $\varepsilon$ enters: one per snap when coverage is converted from $d_\theta$ into $d^*$, and a third when the finished edge is converted back into the model's units, which is the value that pruning tests.

Two further facts make this route usable, and both are established below: its edges must survive pruning, which is what the threshold condition $\kappa\delta\ge3\delta+3\varepsilon$ gives (Lemma~\ref{lem:route}), and its two endpoints must be candidates, which is what coverage of $s_0$ and $s_g$ gives. Equation~\ref{eq:snap} also shows why a landmark must be near its waypoint in both directions: $c_i$ enters through $d^*(w_i,c_i)$ in the edge that ends at it and through $d^*(c_i,w_i)$ in the next one. A third case, treated in Theorem~\ref{thm:approx}(iii), covers pairs that share their nearest landmark, where routing through the graph would only be a detour and the estimator returns $d_\theta(s_0,s_g)$ directly.

The route just described also survives pruning, and its total cost is bounded: this is Lemma~\ref{lem:route} in Appendix~\ref{app:proofs}, which supplies the landmarks $c_0,\dots,c_M$ used by Theorem~\ref{thm:approx}(ii).

\paragraph{The edge threshold.}
The threshold enters only through Lemma~\ref{lem:route}. The constructed route may require an edge containing two snaps and one waypoint gap, with model length up to $3\delta+3\varepsilon$. Dividing by $\delta$ gives
\begin{equation}
\label{eq:constants}
    \kappa \;\ge\; 3 + 3\varepsilon/\delta .
\end{equation}
Assumption~\ref{ass:local} gives $\varepsilon\le\delta$. Thus $\kappa=6$ always satisfies Equation~\ref{eq:constants}, while $\kappa=4$ requires $\varepsilon\le\delta/3$. Increasing the threshold is not free---$\kappa\delta$ is also the radius over which local accuracy must hold---so a larger $\kappa$ admits more edges and requires accuracy over a wider range.

Equation~\ref{eq:constants} is sufficient, not necessary. It ensures graph connectivity by constructing a conservative landmark route along a shortest $d^*$-path. Each edge along the constructed route may include a waypoint gap and two snaps. The graph can contain a shorter landmark chain, for example when landmarks lie densely near the reference path. Such a chain can survive a stricter threshold. The condition is therefore a worst-case guarantee. In practice, smaller values of $\kappa$ may work.

\subsection{Two-Sided Error Bound}
\label{sec:error-bound}

\begin{theorem}[Two-sided hierarchical bound]
\label{thm:approx}
Let Assumptions~\ref{ass:local} and~\ref{ass:coverage} hold with $\kappa$ satisfying Equation~\ref{eq:constants}, and let $\hat{d}_H$ use the untruncated candidate sets $\mathcal{C}_s, \mathcal{C}_g$.  Fix a query pair $(s_0, s_g)$ and put $M = \lceil d^*(s_0, s_g)/\delta \rceil$.
\begin{enumerate}
    \item[(i)] \emph{(No underestimation.)}  In the multi-hop regime, if the route attaining $\hat{d}_H(s_0, s_g)$ uses $m$ graph edges, then
    \[
        \hat{d}_H(s_0, s_g) \;\ge\; d^*(s_0, s_g) - (m + 2)\,\varepsilon .
    \]
    \item[(ii)] \emph{(Bounded overestimation.)}  In the multi-hop regime,
    \[
        \hat{d}_H(s_0, s_g) \;\le\; d^*(s_0, s_g) + M\,(2\delta + 3\varepsilon) + 2\delta .
    \]
    \item[(iii)] \emph{(Same-landmark regime.)}  If $c_s^* = c_g^*$, then $|\hat{d}_H(s_0, s_g) - d^*(s_0, s_g)| \le \varepsilon$.
\end{enumerate}
\end{theorem}

\noindent The proof is given in Appendix~\ref{app:proofs}.


The same two bounds also order two queries whose target distances are far enough apart; Corollary~\ref{cor:rankings} in Appendix~\ref{app:proofs} states this and proves it.

The condition in Corollary~\ref{cor:rankings} applies to sufficiently separated pairs: their target distances must differ by more than the cost $M(2\delta+3\varepsilon)+2\delta$ that the comparison route pays on top of $d^*(s_0,s_g)$, plus the error $(m'+2)\varepsilon$ accumulated on the other query. It does not guarantee the ordering of arbitrarily close pairs or directly imply a Spearman correlation. Instead, it identifies one type of long-range ordering that the hierarchy preserves even when a base model saturates.


The constants are conservative. The proof charges every local error at its worst value, and prices one particular route rather than the route selected by minimization. The result should therefore be read as an existence guarantee for bounded distortion, not as a prediction of the error the estimator will attain.



\paragraph{Beyond MAD.}
This analysis is not specific to MAD, and in fact applies to any quasimetric target for which the base model is locally accurate. Stochastic control introduces an additional issue: an edge may be executed by a local policy that fails to reach its intended target. Appendix~\ref{app:beyond-mad} discusses an options-style abstraction that models these outcomes explicitly.

\begin{figure}[t]
    \centering
    \includegraphics[width=0.78\linewidth]{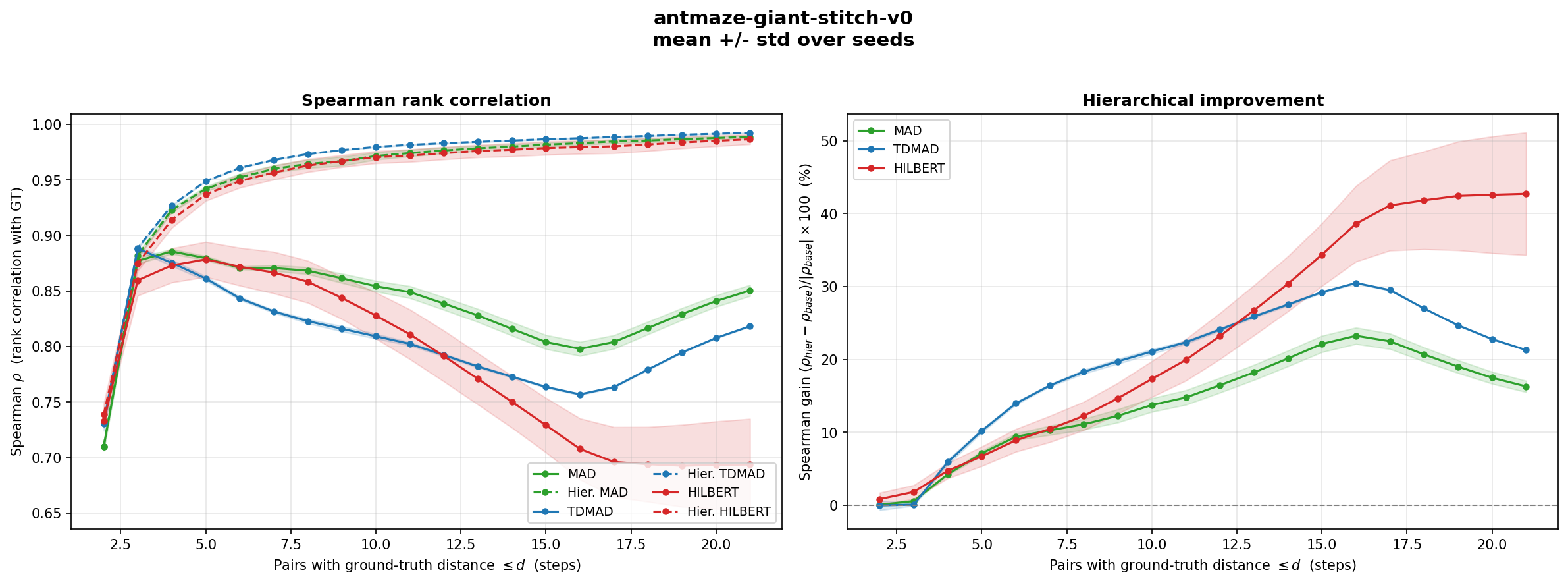}
    \caption{Spearman correlation with grid-maze distance as the evaluated range increases (left) and relative improvement from the hierarchy (right), on the largest setting: AntMaze Giant, \emph{stitch} split. Results show the mean $\pm$ standard deviation over five seeds for all three base models.}
    \label{fig:sp_giant_stitch}
    \vspace{-1em}
\end{figure}

\begin{figure}[t]
    \centering
    \includegraphics[width=0.5\linewidth]{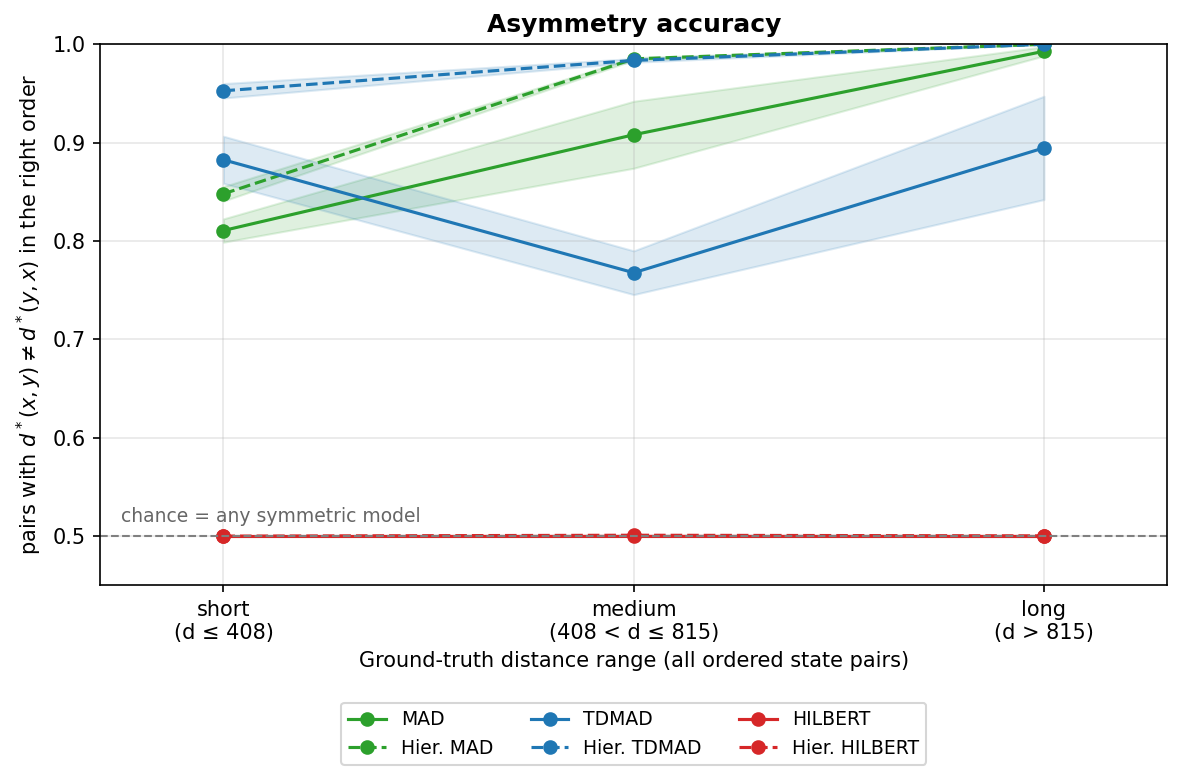}
    \caption{Directional ordering accuracy on cyclic Towers of Hanoi with seven disks. For state pairs $(x,y)$ with $d^*(x,y)\neq d^*(y,x)$, we measure whether each model correctly identifies which direction has the higher distance. We group pairs by $d^*(x,y)$ into short, medium, and long ranges. Solid lines represent base models, and dashed lines represent their hierarchical counterparts. HILBERT's symmetric estimates yield a score of $0.5$, equivalent to chance. We report the mean $\pm$ standard deviation over five seeds.}
    \label{fig:hanoi-asym}
    \vspace{-1.5em}
\end{figure}

\section{Empirical Analysis}
\label{sec:results}

In this section, we present an empirical evaluation of hierarchical distance decomposition. Our analysis is directed by the following questions: Does hierarchical distance decomposition improve the long-horizon MAD estimates produced by existing distance-learning methods? If so, do these distance estimation improvements translate into better downstream planning performance?

\paragraph{Environments and data.}
We evaluate on the PointMaze and AntMaze environments from the OGBench suite~\citep{park2025ogbench}, using the \emph{medium}, \emph{large}, and \emph{giant} environment sizes. For both environments, we evaluate performance using the \emph{navigate} and \emph{stitch} dataset variants. For AntMaze Medium, we also use the \emph{explore} dataset. We also use the cyclic Towers of Hanoi with seven disks~\citep{atkinson1981cyclic}, a discrete domain where the exact MAD is known for every ordered pair of states and is strongly asymmetric (Appendix~\ref{app:hanoi}).


\paragraph{Evaluation metrics.}
To evaluate distance quality, we follow the methodology of \citet{steccanella2026} and compare learned distances with the grid-geodesic distance through the underlying maze. This is computed by applying the Floyd--Warshall algorithm to the maze topology. We measure the agreement between the predicted and ground-truth grid-geodesic distances using Spearman rank correlation. In the tabular Towers of Hanoi environment, we compare against the true ground-truth MAD instead, and report how often each model recovers the true direction of the asymmetry (i.e., which of $d^*(x,y)$ and $d^*(y,x)$ is greater).

While the grid-geodesic distance provides a useful proxy for navigation progress, it is not the exact MAD---particularly in environments like AntMaze where maze topology alone does not capture the agent's embodied dynamics and control effort.  We therefore complement this distance-quality evaluation with planning success rate, which measures whether improvements in the learned distance estimates actually translate into better downstream goal-reaching performance.

\paragraph{Base distances.}
We evaluate hierarchical distance decomposition using three learned base distance models. We use two quasimetric models trained as described in \cite{steccanella2026}: \textbf{MadDist} trained using the contrastive and trajectory-distance losses, and \textbf{TDMAD}, its temporal-difference variant. We also include the \textbf{HILBERT} method of \citet{park2024foundation}, a symmetric distance embedding trained using hindsight relabelling. We train each model once for each environment--dataset combination and seed. We report all the hyperparameters selected for training these base models in Appendix \ref{app:base-models}. Note that the proposed approach is agnostic to the base model used
as long as it does not violate Assumptions \ref{ass:local} and \ref{ass:coverage}.

\paragraph{Hierarchy.}
We construct the hierarchical decomposition separately for each base distance model. Each decomposition uses the corresponding base model for local distance evaluation, so each flat--hierarchical comparison uses \textit{exactly} the same underlying base distance model.

We construct the landmark net using the offline dataset of trajectories provided for each OGbench environment \citep{park2025ogbench}. We use spacing values $\delta$ of $6$ and $4$, a threshold multiplier $\kappa=3$ on Towers of Hanoi and $\kappa=2$ on all other environments, and $k=5$ entry and exit candidates. Appendix~\ref{app:abl-delta-kappa} reports an ablation study over these parameters. We use the max selection rule from Proposition~\ref{prop:covering} and precompute all-pairs shortest paths once for each environment--dataset setting, base distance model, and seed.

\paragraph{Planner.}
To evaluate downstream planning performance, we use a sampling-based planner that considers candidate short-horizon action sequences and uses the learned distance model to select the one that leads to progress towards a given goal. Within each environment, we keep the planner and its search budget fixed across all distance models, varying only the distance function provided to the planner. Full details can be found in Appendix~\ref{app:Planning}.


\paragraph{Results.}

Across all environments, datasets, and base models, the proposed hierarchical distance decomposition improves Spearman correlation compared to each base distance model on its own. Figure~\ref{fig:sp_giant_stitch} shows learning curves for the largest setting, AntMaze Giant on the \emph{stitch} split, and Appendix~\ref{app:distance-results} reports the same comparison for all remaining environment--dataset settings. Across environments, the base models typically lose rank agreement at longer distances, while the hierarchical estimates consistently improve over the base models. We also report hierarchical gains showing that the percentage improvement in how well hierarchical predictions correlate with the reference grid-geodesic distances increases as the distance between state pairs increases.

The same pattern is visible in Table~\ref{tab:planning}, which shows planning success rates. Gains are largest on long-horizon tasks where the base models fail entirely. Several flat baselines achieve near-zero success on giant mazes while hierarchical versions succeed reliably. Gains shrink when the base is already accurate. The hierarchy improves HILBERT on PointMaze, but less so on the hardest AntMaze tasks, where its local predictions are too inaccurate to compose effectively over long horizons.

On cyclic Towers of Hanoi, where we investigate the impact of asymmetry in the true MAD, we observe similar improvements in long-range distance estimation (Appendix~\ref{app:hanoi}). The hierarchical MAD and TDMAD models also successfully recover the directional ordering of distances between states (Figure~\ref{fig:hanoi-asym}). As expected, neither the base nor hierarchical HILBERT models can distinguish between directions because their distance estimates are symmetric, leading to significant inaccuracies in this highly-directed environment.

The patterns in these results align well with the motivation for hierarchical distance decomposition: replacing long-range predictions with compositions of short-range predictions increases their accuracy, and these improvements lead to increased performance on downstream planning tasks.


\begin{table}[t]
  \centering
  \caption{Planning success rate on the \emph{giant} mazes, over 50 episodes, reported as mean $\pm$ standard deviation over five seeds. \emph{Base} is the learned distance used by the planner and \emph{Data} is its dataset type. Flat uses the base distance directly; Hierarchical applies Algorithm~\ref{alg:hierarchical}. The complete results, including the medium and large mazes, are given in Appendix~\ref{app:planning-results}.}
  \label{tab:planning}
  \small
  \setlength{\tabcolsep}{4pt}
  \renewcommand{\arraystretch}{0.95}
  \begin{tabular}{llrrr}
    \toprule
    Environment & Base / Data & Flat & Hierarchical & $\Delta$ \\
    \midrule
    PointMaze Giant & HILBERT / navigate & $0.000 \pm 0.000$ & $0.992 \pm 0.016$ & $+0.992$ \\
     & HILBERT / stitch & $0.000 \pm 0.000$ & $0.984 \pm 0.023$ & $+0.984$ \\
     & MAD / navigate & $0.072 \pm 0.068$ & $0.984 \pm 0.023$ & $+0.912$ \\
     & MAD / stitch & $0.040 \pm 0.062$ & $1.000 \pm 0.000$ & $+0.960$ \\
     & TDMAD / navigate & $0.028 \pm 0.039$ & $1.000 \pm 0.000$ & $+0.972$ \\
     & TDMAD / stitch & $0.000 \pm 0.000$ & $1.000 \pm 0.000$ & $+1.000$ \\
    \midrule
    AntMaze Giant & HILBERT / navigate & $0.000 \pm 0.000$ & $0.540 \pm 0.092$ & $+0.540$ \\
     & HILBERT / stitch & $0.000 \pm 0.000$ & $0.388 \pm 0.057$ & $+0.388$ \\
     & MAD / navigate & $0.004 \pm 0.008$ & $0.992 \pm 0.016$ & $+0.988$ \\
     & MAD / stitch & $0.016 \pm 0.015$ & $0.964 \pm 0.023$ & $+0.948$ \\
     & TDMAD / navigate & $0.000 \pm 0.000$ & $0.980 \pm 0.013$ & $+0.980$ \\
     & TDMAD / stitch & $0.000 \pm 0.000$ & $0.984 \pm 0.015$ & $+0.984$ \\
    \bottomrule
  \end{tabular}
  \vspace{-0.5cm}
\end{table}

\section{Conclusion}
\label{sec:conclusion}

In this paper, we introduced hierarchical distance decomposition, a training-free approach that improves the long-range distance estimates of existing learning distance models by composing a series of shorter-range estimates to answer longer-range queries.

Our empirical evaluation across multiple goal-conditioned tasks demonstrates that the hierarchy improves rank correlation with the reference grid-geodesic distances. Importantly, these improved distance estimates lead to substantial performance gains on downstream planning tasks. The improvements are largest for base models that struggle with long-range distance estimation, and in larger environments where longer-distance queries may be made.

One limitation of the proposed approach is that both the theoretical guarantees and the empirical performance depend on a base model with sufficient local accuracy, as well as adequate landmark coverage between state pairs whose distance might be queried.

An important direction for future work is extending the proposed approach to support online adaptation as the agent explores its environment. Existing distance-learning methods, such as METRA~\citep{park2024metra} and TDMAD~\citep{steccanella2026}, can update the base distance model using new experience. In parallel, the agent could extend its landmark graph to cover newly explored regions, periodically updating its edges, weights, and shortest-path distances to reflect both these new landmarks and changes to the base distance model.

Finally, we believe the proposed method can serve as a foundation for future work exploring \emph{task division}, where local distance models are trained to be accurate only in specific local regions, with long-horizon distance estimation delegated entirely to hierarchical distance decomposition.

\subsection*{AI use statement}
In this work, we used generative AI tools as assistants for writing the \LaTeX{} version of the paper, generating TikZ plots, and developing the code required for the experiments.
We carefully reviewed all AI-assisted content and take full responsibility for the final content of this work, including all text, claims, and artifacts produced with the assistance of generative AI.

\subsection*{Ethics statement}
This work studies distance estimation on standard offline-navigation benchmarks. It involves no human subjects, personally identifiable data, or deployment on physical systems. 

\subsection*{Reproducibility statement}
Algorithm~\ref{alg:hierarchical} specifies the method, while Appendix \ref{app:abl-delta-kappa} and Appendix \ref{app:base-models} specify all the hyperparameters used to construct the hierarchy and train the base models. Based on the acceptance of the paper, we plan to release the full code to reproduce all our results. 

\bibliography{bibliography}

@preamble{ "\providecommand{\noopsort}[1]{}" }

@inproceedings{park2025ogbench,
	title={{OGB}ench: Benchmarking offline goal-conditioned {RL}},
	author={Park, Seohong and Frans, Kevin and Eysenbach, Benjamin and Levine, Sergey},
 	booktitle={Proceedings of the 13th International Conference on Learning Representations},
  	volume={2025},
  	pages={94937--94982},
  	year={2025}
}

@inproceedings{steccanella2026,
  title={Learning the minimum action distance},
  author={Steccanella, Lorenzo and Evans, Joshua B. and {\c{S}}im{\c{s}}ek, {\"O}zg{\"u}r and Jonsson, Anders},
  booktitle={Proceedings of the 43rd International Conference on Machine Learning},
  year={2026},
  publisher={PMLR}
}

@inproceedings{park2024metra,
  title={{METRA}: scalable unsupervised {RL} with metric-aware abstraction},
  author={Park, Seohong and {\noopsort{a}}Rybkin, Oleh and Levine, Sergey},
  booktitle={Proceedings of the 12th International Conference on Learning Representations},
  year={2024}
}

@article{park2024foundation,
      title={Foundation policies with {H}ilbert representations}, 
      author={Seohong Park and {\noopsort{b}}Tobias Kreiman and Sergey Levine},
      year={2024},
      booktitle={Proceedings of the 41st International Conference on Machine Learning},
      pages={39737--39761},
      publisher={PMLR},
}

@inproceedings{wang2023optimal,
  title={Optimal goal-reaching reinforcement learning via quasimetric learning},
  author={Wang, Tongzhou and Torralba, Antonio and Isola, Phillip and Zhang, Amy},
  booktitle={Proceedings of the 40th International Conference on Machine Learning},
  pages={36411--36430},
  year={2023},
  publisher={PMLR}
}

@article{kaelbling1993,
    author = {Kaelbling, Leslie Pack},
    title = {Learning to achieve goals},
    journal = {International Joint Conference on Artificial Intelligence},
    year = {1993},
    pages = {1094--1098},
    volume = {2},
}

@inproceedings{hartikainen2020,
  title={Dynamical distance learning for semi-supervised and unsupervised skill discovery},
  author={Hartikainen, Kristian and Geng, Xinyang and Haarnoja, Tuomas and Levine, Sergey},
  year={2020},
  booktitle={Proceedings of the 8th International Conference on Learning Representations}
}

@inproceedings{steccanella2022,
  title={State representation learning for goal-conditioned reinforcement learning},
  author={Steccanella, Lorenzo and Jonsson, Anders},
  booktitle={Joint European Conference on Machine Learning and Knowledge Discovery in Databases},
  pages={84--99},
  year={2022},
  publisher={Springer}
}

@inproceedings{zheng2026,
  title={Multistep quasimetric learning for scalable goal-conditioned reinforcement learning},
  author={Zheng, Bill Chunyuan and Myers, Vivek and Eysenbach, Benjamin and Levine, Sergey},
  booktitle={Proceedings of the 14th International Conference on Learning Representations},
  year={2026}
}

@inproceedings{myers2025,
  title={Offline goal-conditioned reinforcement learning with quasimetric representations},
  author={Myers, Vivek and Zheng, Bill and Eysenbach, Benjamin and Levine, Sergey},
  journal={Advances in Neural Information Processing Systems},
  volume={38},
  pages={19654--19679},
  year={2026},
  publisher={Curran Associates, Inc.}
}

@article{eysenbach2019,
  title={Search on the replay buffer: Bridging planning and reinforcement learning},
  author={Eysenbach, Ben and Salakhutdinov, Russ R. and Levine, Sergey},
  journal={Advances in Neural Information Processing Systems},
  volume={32},
  year={2019},
  publisher={Curran Associates, Inc.}
}

@inproceedings{baek2025,
  title={Graph-assisted stitching for offline hierarchical reinforcement learning},
  author={Baek, Seungho and Park, Taegeon and Park, Jongchan and Oh, Seungjun and Kim, Yusung},
  booktitle={Proceedings of the 42nd International Conference on Machine Learning},
  pages={2391--2408},
  year={2025},
  publisher={PMLR}
}

@inproceedings{opryshko2026testtime,
  title={Test-Time Graph Search for Goal-Conditioned Reinforcement Learning},
  author={Evgenii Opryshko and Junwei Quan and Claas A Voelcker and Yilun Du and Igor Gilitschenski},
  booktitle={Proceedings of the 43rd International Conference on Machine Learning},
  year={2026},
  publisher={PMLR}
}

@inproceedings{kobanda2026,
  title={Adaptive Quasimetric Mapping: Principled Topological Abstraction for Robust Offline Goal-Conditioned Navigation},
  author={Anthony Kobanda and Waris Radji and Odalric-Ambrym Maillard and R{\'e}my Portelas},
  booktitle={Proceedings of the 43rd International Conference on Machine Learning},
  year={2026},
  publisher={PMLR}
}

@inproceedings{yang2020,
  title = 	 {Plan2Vec: Unsupervised Representation Learning by Latent Plans},
  author =       {Yang, Ge and Zhang, Amy and Morcos, Ari and Pineau, Joelle and Abbeel, Pieter and Calandra, Roberto},
  booktitle = 	 {Proceedings of the 2nd Conference on Learning for Dynamics and Control},
  pages = 	 {935--946},
  year = 	 {2020},
  publisher =    {PMLR},
}

@inproceedings{bagatella2023,
 title={Goal-conditioned Offline Planning from Curious Exploration},
 author = {Bagatella, Marco and Martius, Georg},
 booktitle = {Advances in Neural Information Processing Systems},
 volume = {36},
 pages = {15358--15383},
 publisher = {Curran Associates, Inc.},
 year = {2023}
}

@inproceedings{park2023,
  title={{HIQL}: offline goal-conditioned {RL} with latent states as actions},
  author={Park, Seohong and Ghosh, Dibya and Eysenbach, Benjamin and Levine, Sergey},
  booktitle={Advances in Neural Information Processing Systems},
  volume={36},
  pages={34866--34891},
  year={2023},
  publisher = {Curran Associates, Inc.},
}

@inproceedings{canesse2025,
  title={Navigation with {QPHIL}: Quantizing Planner for Hierarchical Implicit {Q}-Learning},
  author={Alexi Canesse and Mathieu Petitbois and Ludovic Denoyer and Sylvain Lamprier and R{\'e}my Portelas},
  booktitle={Proceedings of the 2025 International Joint Conference on Neural Networks},
  year={2025},
  publisher={IEEE}
}

@inproceedings{wibault2026,
  title={Abstraction for Offline Goal-Conditioned Reinforcement Learning},
  author={Clarisse Wibault and Alexander D. Goldie and Antonio L. Villares and Maike A. Osborne and Jakob N. Foerster},
  booktitle={Workshop on Decision-Making from Offline Datasets to Online Adaptation: Black-Box Optimization to Reinforcement Learning},
  year={2026},
}

@article{ahn2026,
  title={Option-aware temporally abstracted value for offline goal-conditioned reinforcement learning},
  author={Ahn, Hongjoon and Choi, Heewoong and Han, Jisu and Moon, Taesup},
  journal={Advances in Neural Information Processing Systems},
  volume={38},
  year={2026},
  publisher={Curran Associates, Inc.}
}

@inproceedings{rodriguezsanchez2024,
  title={Learning Abstract World Models for Value-preserving Planning with Options},
  author={Rafael Rodriguez-Sanchez and George Konidaris},
  booktitle={Proceedings of the 1st Reinforcement Learning Conference},
  year={2024},
  publisher={RLJ}
}

@article{piriyakulkij2026,
  title={Joint Learning of Hierarchical Neural Options and Abstract World Model},
  author={Piriyakulkij, Wasu T. and Lehrach, Wolfgang and Ellis, Kevin and Murphy, Kevin},
  journal={arXiv preprint arXiv:2602.02799},
  year={2026}
}

@inproceedings{kingma2015adam,
  title={Adam: A Method for Stochastic Optimization},
  author={Kingma, Diederik P. and Ba, Jimmy},
  booktitle={Proceedings of the 3rd International Conference on Learning Representations},
  year={2015},
  publisher={PMLR}
}

@article{klambauer2017self,
  title={Self-normalizing neural networks},
  author={Klambauer, G{\"u}nter and Unterthiner, Thomas and Mayr, Andreas and Hochreiter, Sepp},
  journal={Advances in neural information processing systems},
  volume={30},
  year={2017}
}

@inproceedings{andrychowicz2017her,
  title={{Hindsight Experience Replay}},
  author={Andrychowicz, Marcin and Wolski, Filip and Ray, Alex and Schneider, Jonas and Fong, Rachel and Welinder, Peter and McGrew, Bob and Tobin, Josh and Abbeel, Pieter and Zaremba, Wojciech},
  booktitle={Advances in Neural Information Processing Systems},
  volume={30},
  year={2017},
  publisher = {Curran Associates, Inc.},
}

@article{newey1987expectile,
  title={{Asymmetric Least Squares Estimation and Testing}},
  author={Newey, Whitney K and Powell, James L},
  journal={Econometrica: Journal of the Econometric Society},
  pages={819--847},
  year={1987},
}

@inproceedings{kostrikov2022iql,
  title={{Offline Reinforcement Learning with Implicit Q-Learning}},
  author={Kostrikov, Ilya and Nair, Ashvin and Levine, Sergey},
  year={2022},
  booktitle={Proceedings of the 10th International Conference on Learning Representations}
}

@article{mnih2015dqn,
  title={Human-level control through deep reinforcement learning},
  author={Mnih, Volodymyr and Kavukcuoglu, Koray and Silver, David and Rusu, Andrei A and Veness, Joel and Bellemare, Marc G and Graves, Alex and Riedmiller, Martin and Fidjeland, Andreas K and Ostrovski, Georg and others},
  journal={nature},
  volume={518},
  number={7540},
  pages={529--533},
  year={2015},
  publisher={Nature Publishing Group}
}

@inproceedings{van2016doubledqn,
  title={{Deep Reinforcement Learning with Double Q-Learning}},
  author={Van Hasselt, Hado and Guez, Arthur and Silver, David},
  booktitle={Proceedings of the AAAI conference on artificial intelligence},
  volume={30},
  year={2016}
}

@article{hendrycks2016gelu,
  title={{G}aussian {E}rror {L}inear {U}nits ({GELU}s)},
  author={Hendrycks, Dan and Gimpel, Kevin},
  journal={arXiv preprint arXiv:1606.08415},
  year={2016}
}

@article{atkinson1981cyclic,
  title={The cyclic towers of Hanoi},
  author={Atkinson, MD},
  journal={Information Processing Letters},
  volume={13},
  number={3},
  pages={118--119},
  year={1981},
  publisher={Elsevier}
}
\bibliographystyle{iclr2027_conference}

\clearpage
\appendix

\begin{center}
    {\LARGE \textsc{Appendix}}
\end{center}
\vspace{1.5em}

\section{Proofs}
\label{app:proofs}

In this appendix we present the proofs deferred from Section~\ref{sec:theory}.

\begin{lemma}[Only local evaluations are used]
\label{lem:all-local}
Every evaluation of $d_\theta$ that enters $\hat{d}_H(s_0, s_g)$ in the multi-hop regime---edge weights and legs alike---is performed on a pair $(x, y)$ such that $d_\theta(x, y) \le \kappa\delta$, and hence satisfies $|d_\theta(x, y) - d^*(x, y)| \le \varepsilon$.
\end{lemma}
\begin{proof}
An edge $(c_i,c_j)$ is retained only if $d_\theta(c_i,c_j)\le\kappa\delta$. Likewise, $\mathcal{C}_s$ contains only landmarks with $d_\theta(s_0,c)\le\kappa\delta$, and $\mathcal{C}_g$ contains only landmarks with $d_\theta(c,s_g)\le\kappa\delta$. Assumption~\ref{ass:local} therefore applies through its $d_\theta$ condition.
\end{proof}

\begin{lemma}[A short reference route survives pruning]
\label{lem:route}
Let Assumptions~\ref{ass:local} and~\ref{ass:coverage} hold, suppose the pruning threshold satisfies $\kappa\delta \ge 3\delta + 3\varepsilon$, and let $(s_0, s_g)$ be a query pair with $M = \lceil d^*(s_0, s_g)/\delta \rceil$. Then there are landmarks $c_0, \dots, c_M \in \mathcal{C}$ such that $d_\theta(s_0, c_0) \le \delta$ and $d_\theta(c_M, s_g) \le \delta$, consecutive distinct landmarks are joined by edges of $G_\mathcal{C}$, and
\[
    \sum_{i=1}^{M} d_\theta(c_{i-1}, c_i) \;\le\; d^*(s_0, s_g) + M(2\delta + 3\varepsilon) .
\]
\end{lemma}

\begin{proof}
Assumption~\ref{ass:coverage} gives a shortest $d^*$-path from $s_0$ to $s_g$ whose states are all covered. The path is cut into $M$ pieces of length at most $\delta$, with the form $w_0=s_0,w_1,\dots,w_{M-1},w_M=s_g$. The pieces are ordered along a shortest path, such that $d^*(w_{i-1},w_i)\le\delta$ and $\sum_{i=1}^{M} d^*(w_{i-1},w_i)=d^*(s_0,s_g)$.

Let $c_i$ be a landmark covering $w_i$, so that $\max\bigl(d_\theta(w_i,c_i),d_\theta(c_i,w_i)\bigr)\le\delta$; the cases $i=0$ and $i=M$ are the two endpoint bounds. Coverage bounds each snap by $\delta$ in the model, within the local range $\kappa\delta$, and are converted by Equation~\ref{eq:no-shortcut} into the target distance: $\max\bigl(d^*(w_i,c_i),d^*(c_i,w_i)\bigr)\le\delta+\varepsilon$. Equation~\ref{eq:snap} then prices the edge $(c_{i-1},c_i)$ as
\[
    d^*(c_{i-1}, c_i)
    \;\le\; (\delta+\varepsilon) + d^*(w_{i-1}, w_i) + (\delta+\varepsilon)
    \;\le\; 3\delta + 2\varepsilon \;\le\; \kappa\delta .
\]
Since $d^*(c_{i-1},c_i)\le\kappa\delta$, Equation~\ref{eq:survives} applies and converts the edge back into the model's units:
\begin{align}
    d_\theta(c_{i-1}, c_i)
    &\;\le\; d^*(c_{i-1}, c_i) + \varepsilon
    && \text{by Equation~\ref{eq:survives}}
    \nonumber
    \\
    &\;\le\; \bigl[(\delta+\varepsilon) + d^*(w_{i-1}, w_i) + (\delta+\varepsilon)\bigr] + \varepsilon
    \nonumber
    \\
    &\;=\; d^*(w_{i-1}, w_i) + 2\delta + 3\varepsilon
    \label{eq:edge-model-c}
    \\
    &\;\le\; 3\delta + 3\varepsilon \;\le\; \kappa\delta ,
    && \text{since } d^*(w_{i-1}, w_i) \le \delta
    \nonumber
\end{align}
where the last inequality is the hypothesis $\kappa\delta\ge3\delta+3\varepsilon$. The final value is the test applied by pruning, so every consecutive distinct pair $(c_{i-1},c_i)$ is an edge of $G_\mathcal{C}$.

It remains to sum Equation~\ref{eq:edge-model-c} over the $M$ edges:
\begin{align}
    \sum_{i=1}^{M} d_\theta(c_{i-1}, c_i)
    &\;\le\; \sum_{i=1}^{M} \bigl[\, d^*(w_{i-1}, w_i) + 2\delta + 3\varepsilon \,\bigr]
    && \text{by Equation~\ref{eq:edge-model-c}}
    \nonumber
    \\
    &\;=\; \sum_{i=1}^{M} d^*(w_{i-1}, w_i) \;+\; M(2\delta + 3\varepsilon)
    \nonumber
    \\
    &\;=\; d^*(s_0, s_g) + M(2\delta + 3\varepsilon) ,
    \nonumber
\end{align}
where the last line uses $\sum_{i=1}^{M} d^*(w_{i-1},w_i)=d^*(s_0,s_g)$, established in the first paragraph. This is the claimed bound.
\end{proof}

With the two lemmas in place we restate the theorem and prove it.

\addtocounter{theorem}{-1}
\begin{theorem}[Two-sided hierarchical bound, restated]
Let Assumptions~\ref{ass:local} and~\ref{ass:coverage} hold with $\kappa$ satisfying Equation~\ref{eq:constants}, and let $\hat{d}_H$ use the untruncated candidate sets $\mathcal{C}_s, \mathcal{C}_g$.  Fix a query pair $(s_0, s_g)$ and put $M = \lceil d^*(s_0, s_g)/\delta \rceil$.
\begin{enumerate}
    \item[(i)] \emph{(No underestimation.)}  In the multi-hop regime, if the route attaining $\hat{d}_H(s_0, s_g)$ uses $m$ graph edges, then
    \[
        \hat{d}_H(s_0, s_g) \;\ge\; d^*(s_0, s_g) - (m + 2)\,\varepsilon .
    \]
    \item[(ii)] \emph{(Bounded overestimation.)}  In the multi-hop regime,
    \[
        \hat{d}_H(s_0, s_g) \;\le\; d^*(s_0, s_g) + M\,(2\delta + 3\varepsilon) + 2\delta .
    \]
    \item[(iii)] \emph{(Same-landmark regime.)}  If $c_s^* = c_g^*$, then $|\hat{d}_H(s_0, s_g) - d^*(s_0, s_g)| \le \varepsilon$.
\end{enumerate}
\end{theorem}

\paragraph{Proof of Theorem~\ref{thm:approx}.}
\emph{(i)} Let $s_0\to c_0\to c_1\to\dots\to c_m\to s_g$ attain the minimum in Equation~\ref{eq:hier_estimate}. Then
$\hat{d}_H(s_0,s_g)=d_\theta(s_0,c_0)+\sum_{i=1}^{m}d_\theta(c_{i-1},c_i)+d_\theta(c_m,s_g)$.
By Lemma~\ref{lem:all-local}, every one of these $m+2$ terms is at least the target length of its segment, minus $\varepsilon$. The segments form a path from $s_0$ to $s_g$, so by the triangle inequality their target lengths sum to at least $d^*(s_0,s_g)$. If no route exists, $\hat d_H(s_0,s_g)=\infty$ and the claim is immediate.

\emph{(ii)} The estimate is the minimum over feasible routes, so the route $s_0\to c_0\to\dots\to c_M\to s_g$ constructed in Lemma~\ref{lem:route} upper-bounds it. That route is feasible: its endpoints satisfy $c_0\in\mathcal{C}_s$ and $c_M\in\mathcal{C}_g$, and its graph edges bound $D_h(c_0,c_M)$. Its two legs satisfy $d_\theta(s_0,c_0)\le\delta$ and $d_\theta(c_M,s_g)\le\delta$, also by Lemma~\ref{lem:route}. Hence
\begin{align}
    \hat{d}_H(s_0, s_g)
    &\;\le\; d_\theta(s_0, c_0) \;+\; \sum_{i=1}^{M} d_\theta(c_{i-1}, c_i) \;+\; d_\theta(c_M, s_g)
    && \text{minimality over routes}
    \nonumber
    \\
    &\;\le\; \delta \;+\; \Bigl[\, d^*(s_0, s_g) + M(2\delta + 3\varepsilon) \,\Bigr] \;+\; \delta
    && \text{legs; Lemma~\ref{lem:route}}
    \nonumber
    \\
    &\;=\; d^*(s_0, s_g) + M(2\delta + 3\varepsilon) + 2\delta ,
    \nonumber
\end{align}
which is the claim. Each term is accounted for: $d^*(s_0,s_g)$ is the progress along the reference path, $2M\delta$ are the two snaps of each of the $M$ edges, $3M\varepsilon$ are the three conversions between $d_\theta$ and $d^*$ charged per edge, and $2\delta$ are the entry and exit legs.

\emph{(iii)} Let $c=c_s^*=c_g^*$. The covering landmark $c_0$ of $s_0$ satisfies $d_\theta(s_0,c_0)\le\delta$. Since $c_s^*$ minimizes $d_\theta(s_0,\cdot)$ over $\mathcal C$, we have $d_\theta(s_0,c)\le\delta$. Similarly, $d_\theta(c,s_g)\le\delta$. Equation~\ref{eq:no-shortcut} gives $d^*(s_0,c)\le\delta+\varepsilon$ and $d^*(c,s_g)\le\delta+\varepsilon$. Therefore $d^*(s_0,s_g)\le2\delta+2\varepsilon\le\kappa\delta$, and Assumption~\ref{ass:local} applies to the returned value $d_\theta(s_0,s_g)$ through its $d^*$ condition. \hfill$\square$

\begin{corollary}[Order preservation]
\label{cor:rankings}
Under the hypotheses of Theorem~\ref{thm:approx}, let $(s_0, s_g)$ and $(s_0', s_g')$ be two multi-hop queries, with $M = \lceil d^*(s_0, s_g)/\delta \rceil$ and $m'$ the number of graph edges on the route attaining $\hat{d}_H(s_0', s_g')$.  Then,
\[
    d^*(s_0', s_g') \;>\; d^*(s_0, s_g) + M\,(2\delta + 3\varepsilon) + 2\delta + (m' + 2)\,\varepsilon
    \quad\Longrightarrow\quad
    \hat{d}_H(s_0', s_g') \;>\; \hat{d}_H(s_0, s_g) .
\]
\end{corollary}
\begin{proof}
By (i) applied to $(s_0',s_g')$ and (ii) applied to $(s_0,s_g)$,
\begin{align}
    \hat{d}_H(s_0', s_g')
    &\;\ge\; d^*(s_0', s_g') - (m' + 2)\,\varepsilon
    && \text{by (i)}
    \nonumber
    \\
    &\;>\; d^*(s_0, s_g) + M\,(2\delta + 3\varepsilon) + 2\delta
    && \text{by the hypothesis}
    \nonumber
    \\
    &\;\ge\; \hat{d}_H(s_0, s_g)
    && \text{by (ii)} .
    \nonumber
\end{align}
\end{proof}

\section{One-Sided Landmark Nets}
\label{app:max-rule}

In Section~\ref{sec:construction} a state $s$ is skipped when some landmark $c$ satisfies $\max\bigl(d_\theta(s,c),d_\theta(c,s)\bigr)\le\delta$, so that one landmark is close to $s$ in both directions. This appendix explains why that requirement is the principled one and compares it with the cheaper alternative of building two one-sided nets: a \emph{forward} net, which skips $s$ when some landmark satisfies $d_\theta(s,c)\le\delta$; a \emph{backward} net, which skips $s$ when some landmark satisfies $d_\theta(c,s)\le\delta$; and their union. The counter example below shows that the two one-sided nets, and their union, can make the estimate infinite even when the base model is exact.

\begin{figure}[!htb]
\centering
\begin{tikzpicture}[
  >=stealth,
  st/.style={circle, draw, fill=white, inner sep=0pt, minimum size=15pt, font=\scriptsize},
  lm/.style={rectangle, draw=orange!60!black, fill=orange!35, inner sep=0pt, minimum size=14pt, font=\scriptsize},
  tr/.style={->, gray!60},
  term/.style={->, very thick, blue!70!black},
  inf/.style={->, very thick, dashed, red!75!black, rounded corners=4pt},
  val/.style={fill=white, inner sep=1pt, font=\scriptsize}]

  \node[lm, minimum size=8pt] (k1) at (0.4, 2.2) {};
  \node[anchor=west, font=\scriptsize, xshift=2pt] (t1) at (k1.east) {selected landmark};
  \node[st, minimum size=8pt, anchor=west, xshift=12pt] (k2) at (t1.east) {};
  \node[anchor=west, font=\scriptsize, xshift=2pt] (t2) at (k2.east) {not selected};
  \draw[tr] ([xshift=12pt]t2.east) -- ++(0.6, 0) coordinate (k3);
  \node[anchor=west, font=\scriptsize, xshift=2pt] (t3) at (k3) {transition};
  \draw[term] ([xshift=12pt]t3.east) -- ++(0.6, 0) coordinate (k4);
  \node[anchor=west, font=\scriptsize, xshift=2pt] at (k4) {term of $\hat d_H$};

  \begin{scope}
    \node[lm] (aq) at (2.375, 1.35) {$q$};
    \node[lm] (ap) at (2.375, -1.35) {$p$};
    \node[lm] (av0) at (0, 0) {$v_0$};
    \foreach \i in {1,...,5} \node[st] (av\i) at (0.95*\i, 0) {$v_{\i}$};
    \foreach \i in {0,...,5} {
      \draw[tr] (aq) -- (av\i);
      \draw[tr] (av\i) -- (ap);
    }
    \foreach \i/\j in {0/1, 1/2, 2/3, 3/4, 4/5} \draw[tr, <->] (av\i) -- (av\j);
    \node[anchor=west, font=\scriptsize] at ([xshift=2pt]aq.east) {initial};
    \node[anchor=west, font=\scriptsize] at ([xshift=2pt]ap.east) {terminal};
    \draw[inf] (av0) to[bend left=40] node[val, text=red!75!black] {$\infty$} (aq);
    \draw[term] (aq) -- node[val, right] {$1$} (av5);
    \node[font=\small, align=center] at (2.375, -2.2) {(a) Union of one-sided nets\\$\hat d_H(v_0,v_5) = 0 + \infty + 1 = \infty$};
  \end{scope}

  \begin{scope}[xshift=6.3cm]
    \node[lm] (bq) at (2.375, 1.35) {$q$};
    \node[lm] (bp) at (2.375, -1.35) {$p$};
    \foreach \i in {0,2,4} \node[lm] (bv\i) at (0.95*\i, 0) {$v_{\i}$};
    \foreach \i in {1,3,5} \node[st] (bv\i) at (0.95*\i, 0) {$v_{\i}$};
    \foreach \i in {0,...,5} {
      \draw[tr] (bq) -- (bv\i);
      \draw[tr] (bv\i) -- (bp);
    }
    \foreach \i/\j in {0/1, 1/2, 2/3, 3/4, 4/5} \draw[tr, <->] (bv\i) -- (bv\j);
    \node[anchor=west, font=\scriptsize] at ([xshift=2pt]bq.east) {initial};
    \node[anchor=west, font=\scriptsize] at ([xshift=2pt]bp.east) {terminal};
    \draw[term] (bv0) to[bend left=50] node[val, above] {$2$} (bv2);
    \draw[term] (bv2) to[bend left=50] node[val, above] {$2$} (bv4);
    \draw[term] (bv4) -- node[val, above] {$1$} (bv5);
    \node[font=\small, align=center] at (2.375, -2.2) {(b) Max rule\\$\hat d_H(v_0,v_5) = 0 + 2 + 2 + 1 = 5$};
  \end{scope}
\end{tikzpicture}
\caption{The example of Appendix~\ref{app:max-rule} for the query $(v_0,v_5)$, with $d^*(v_0,v_5)=5$. Gray arrows are unit-cost transitions and colored arrows are the terms of $\hat d_H$. (a)~The one-sided union keeps $v_0$ as its only corridor landmark, and nothing reaches $q$, so $D_h(v_0,q)=\infty$ (dashed). (b)~The max rule also selects $v_2$ and $v_4$ and is exact.}
\label{fig:one-sided}
\end{figure}
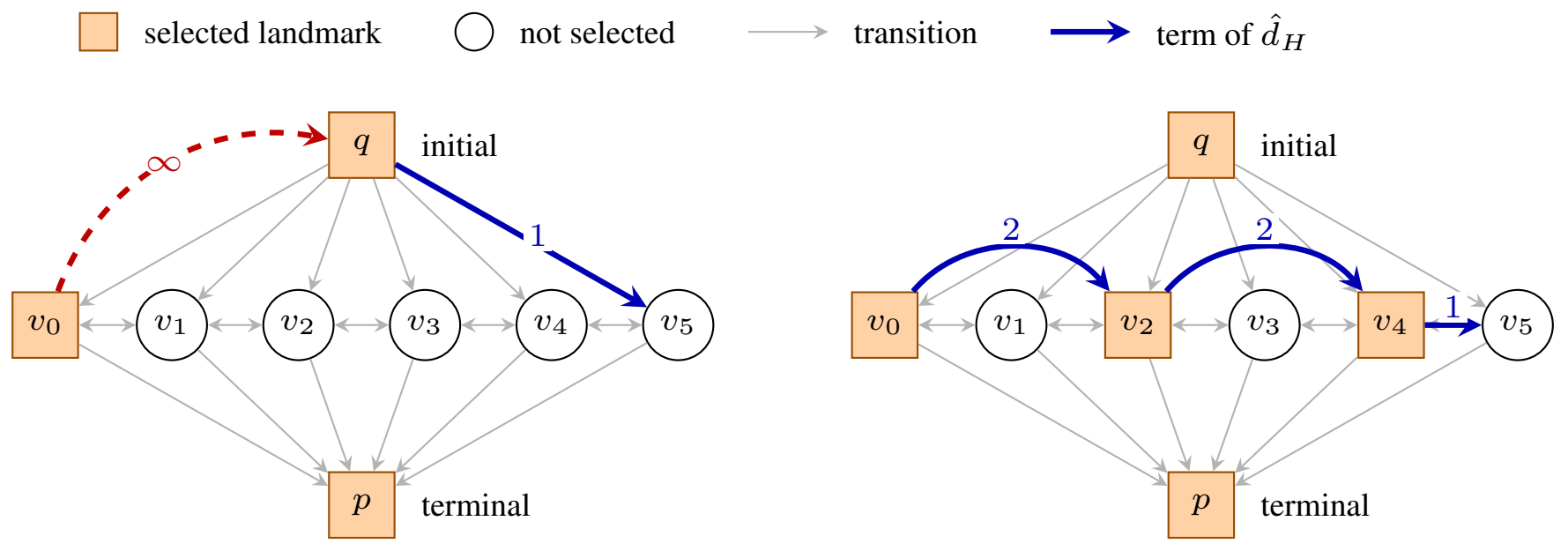

\paragraph{Example.}
Consider a deterministic MDP with unit action costs, and three kinds of states (Figure~\ref{fig:one-sided}): corridor states $v_0,\dots,v_n$ with transitions $v_i\leftrightarrow v_{i+1}$; an initial state $q$ with a transition $q\to v_i$ to every corridor state and no transition into $q$; and a terminal state $p$ with a transition $v_i\to p$ from every corridor state. Every episode starts at $q$, moves to a corridor state, and can terminate only at $p$. For all $i,j$,
\[
    d^*(v_i,v_j)=|i-j|,\qquad d^*(q,v_i)=d^*(v_i,p)=1,\qquad d^*(q,p)=2,
\]
while $d^*(p,y)=\infty$ for every $y\ne p$ and $d^*(y,q)=\infty$ for every $y\ne q$. Let the base model be exact, $d_\theta=d^*$ (so $\varepsilon=0$), and let $n=5$, $\delta=1$, and $\kappa=4$. Consider the dataset of the two episodes
\[
    \tau_1=(q,\,v_0,\,p), \qquad \tau_2=(q,\,v_0,\,v_1,\,v_2,\,v_3,\,v_4,\,v_5,\,p),
\]
which together visit every state. Algorithm~\ref{alg:hierarchical} visits them in this order, and the states of each episode in time order. We consider the query $(v_0,v_5)$, for which $d^*(v_0,v_5)=5$, and Figure~\ref{fig:one-sided} shows the landmarks that each rule selects.

\emph{Backward rule.} Both episodes start at $q$, so $q$ is the first state visited and becomes a landmark. Since $d^*(q,v_i)=1$ for every $i$, the backward rule skips every corridor state, and it selects $p$ because $d^*(q,p)=2$. The backward net is therefore $\{q,p\}$. For two corridor states $v_a$ and $v_b$, the only entry candidate is $p$, because $d^*(v_a,q)=\infty$, and the only exit candidate is $q$, because $d^*(p,v_b)=\infty$. The query is in the multi-hop regime and $D_h(p,q)=\infty$, so $\hat d_H(v_a,v_b)=\infty$ for every pair of corridor states, including neighbours with $d^*(v_a,v_b)=1$.

\emph{Forward rule and the union.} The forward rule selects $q$, then $v_0$ because $d^*(v_0,q)=\infty$, and then $p$ at the end of $\tau_1$ because $d^*(p,y)=\infty$ for every $y\ne p$. From then on $d^*(v_i,p)=1$ covers every corridor state, so $\tau_2$ adds nothing and the forward net, like the union, is $\{q,v_0,p\}$. Since $d^*(v_0,v_5)=5>\kappa\delta$, the only exit candidate for $v_5$ is $q$, and $D_h(c,q)=\infty$ for every landmark $c\ne q$. Hence
\[
    \hat d_H(v_0,v_5)=\infty \qquad\text{while}\qquad d^*(v_0,v_5)=5 .
\]

\emph{Max rule.} The max rule also selects $q$ and $p$, but neither covers a corridor state, because each is at infinite distance from it in one direction. Corridor states are therefore covered only by corridor landmarks, and $v_0$, $v_2$, and $v_4$ are selected. The candidate sum with entry leg $d^*(v_0,v_0)=0$, edges $v_0\to v_2\to v_4$ of length $2\le\kappa\delta$ each, and exit leg $d^*(v_4,v_5)=1$ equals $5$.

\paragraph{The analysis needs both directions.}
The example above exhibits one-way in which the one-sided $\delta$ nets fail. The same condition also enters the analysis of Section~\ref{sec:theory}. The upper bound of Theorem~\ref{thm:approx}(ii) needs one candidate sum close to $d^*(s_0,s_g)$. Take waypoints $w_0=s_0,\dots,w_M=s_g$ along a shortest path from $s_0$ to $s_g$ and a landmark $c_i$ for each $w_i$. The triangle inequality bounds the terms of that sum by
\[
    d^*(c_{i-1},c_i) \;\le\; d^*(c_{i-1},w_{i-1}) + d^*(w_{i-1},w_i) + d^*(w_i,c_i) ,
\]
so an interior landmark $c_i$ appears in two consecutive terms: the term that ends at $c_i$ is controlled by $d^*(w_i,c_i)$, and the term that starts at $c_i$ by $d^*(c_i,w_i)$. The max rule bounds both quantities at the same landmark. The two one-sided nets bound one each, at possibly different landmarks $c$ and $c'$, and chaining them would add a term $d^*(c,c')$ that neither covering condition controls.

\section{Beyond Minimum-Action Distance}
\label{app:beyond-mad}

The analysis in Section~\ref{sec:theory} applies to any quasimetric target with a locally accurate estimator. In a stochastic control problem, however, a graph edge must also be executable. We use \emph{option} to mean the local policy intended to carry the agent from one landmark to another.

Shortest path in $G_\mathcal C$ is appropriate when every edge is implemented by a \emph{proper} option: it reaches its target landmark with probability one, and its edge weight represents expected execution cost. The abstract transition is then deterministic and expected costs compose, so Floyd--Warshall gives the optimal cost of a landmark chain. If an option can fail, terminate at another landmark, or fail to terminate, this interpretation breaks down. The landmark graph is then an abstract MDP, and deterministic shortest path is optimistic because it treats every selected transition as certain. Replanning after each subgoal helps with compounding execution failures but does not correct this optimistic edge model.

One remedy is an options-style semi-Markov abstraction. Represent an option $o_{ij}$, which attempts to reach $c_j$ from $c_i$, by its expected cost $\bar c(o_{ij})$ and termination distribution $P(c'\mid c_i,o_{ij})$, where $c'$ is the landmark closest to the state at which the option stops. Long-range values then solve the abstract stochastic shortest-path problem
\begin{equation}
\label{eq:abstract-ssp}
    V_g(c) \;=\; \min_{o \in \mathcal{O}(c)} \Bigl[\, \bar{c}(o) + \sum_{c' \in \mathcal{C}} P(c' \mid c, o)\, V_g(c') \,\Bigr],
    \qquad V_g(c_g) = 0,
\end{equation}
The query then replaces $D_h(c_s,c_g)$ in Equation~\ref{eq:hier_estimate} with $V_g(c_s)$. If every option is proper, then $P(c'\mid c_i,o_{ij})=\mathbbm{1}[c'=c_j]$ and Equation~\ref{eq:abstract-ssp} reduces to shortest path in $G_\mathcal C$. If an option succeeds with probability $p$ and otherwise returns near its start, the recursion accounts for retries and assigns cost on the order of $\bar c(o_{ij})/p$. Rare-transition shortcuts are therefore penalized by the model rather than filtered through a hand-tuned threshold. The additional cost is estimating an outcome distribution for each option from rollouts and solving one abstract stochastic shortest-path problem per goal landmark.

\section{Base Models}
\label{app:base-models}

In this section, we describe the implementation details of each base model included in our evaluation.

\subsection{Computer Resources}

We run all experiments on a single NVIDIA RTX 4070 GPU with 8GB of VRAM and an Intel i7-4700-HX with 32GB of RAM.

\subsection{MadDist}

To train the MadDist distance models, we used the Adam optimizer with a learning rate of ${1}\times{10^{-4}}$, a batch size of 256 for the objective ($\mathcal{L}_o$, $\mathcal{L}_r$), and a separate batch of size 1024 for the constraint loss ($\mathcal{L}_c$).
For our main experiment, we used the simple quasimetric function \cite{steccanella2026} and a latent dimension size of 512. In contrast with the original code of \cite{steccanella2026} we use $w_r = 1$, $w_c = 0.1$ this enforce the model to be more locally accurate, making the constrain stronger.

The full set of hyperparameter values used to train the MadDist models can be found in Table~\ref{tab:mad_params}.

\begin{table}[ht]
\caption{Hyperparameters used to train the MadDist algorithm.}
\label{tab:mad_params}
\centering
\begin{tabular}{ll}
\hline
\textbf{Hyperparameter}                      & \textbf{Value}                              \\ \hline
Quasimetric Function & $d_{simple}$ \\
Optimizer & Adam~\cite{kingma2015adam} \\
Learning Rate & ${1}\times{10^{-4}}$ \\
Batch Size ($\mathcal{L}_o$, $\mathcal{L}_r$) & 256 \\
Batch Size ($\mathcal{L}_c$) & 1024 \\
Activation Function (Hidden Layers) & SELU \cite{klambauer2017self} \\
Neural Network & (512, 512, 256, 128)\\
$w_r$ & 1\\
$w_c$ & 0.1 \\
$d_{max}$ & 100, 500 \\
$H_c$ & 6 \\
\hline
\end{tabular}
\end{table}

\subsection{TDMadDist}

To train the TDMadDist distance models, we used the the Adam optimizer with a learning rate of ${1}\times{10^{-4}}$, a batch size of 256 for the objective ($\mathcal{L}_o$, $\mathcal{L}_r$), and a separate batch of size 1024 for the constraint loss ($\mathcal{L}_c$). For our main experiment, we used the novel simple quasimetric function and a latent dimension size of 512.

For TDMadDist, we remove the hyperparameter $d_{\max}$ from the MadDist algorithm, because it is not included in TDMadDist's objective ($\mathcal{L}_r$). The temporal-difference update used when training the TDMadDist distance models involves the use of a target network, $d_{\theta'}$, which is updated using a Polyak averaging factor $\tau = 0.005$. 

The full set of hyperparameter values used to train the TDMadDist models can be found in Table~\ref{tab:tdmad_params}.

\begin{table}[ht]
    \caption{Hyperparameters used to train the TDMadDist algorithm.}
    \label{tab:tdmad_params}
    \centering
    \begin{tabular}{ll}
    \hline
    \textbf{Hyperparameter}                      & \textbf{Value}                              \\ \hline
    Quasimetric Function & $d_{simple}$ \\
    Optimizer & Adam~\cite{kingma2015adam} \\
    Learning Rate & ${1}\times{10^{-4}}$ \\
    Batch Size ($\mathcal{L}_o$, $\mathcal{L}_r$) & 256 \\
    Batch Size ($\mathcal{L}_c$) & 1024 \\
Activation Function (Hidden Layers) & SELU \cite{klambauer2017self} \\
    Neural Network & (512, 512, 256, 128)\\
    $w_r$ & 1 \\
    $w_c$ & 0.1 \\
    $H_c$ & 6 \\
    $\tau$ & 0.005 \\

    \hline
    \end{tabular}
    \end{table}

\subsection{Hilbert Representation}

We trained Hilbert representation models following the approach of \citet{park2024foundation}, using action-free Implicit Q-Learning~(IQL)~\citep{park2023} and Hindsight Experience Replay~(HER)~\citep{andrychowicz2017her}.

We used a dataset of state--next-state pairs $(s, s')$, which we relabeled using HER to produce state--next-state--goal tuples $(s, s', g)$. Goals were sampled from a geometric distribution $\text{Geom}(\gamma)$ over future states in the same trajectory with probability $0.625$, and uniformly from the entire dataset with probability $0.375$.

We trained the Hilbert representation model $\phi$ to minimize the temporal-difference loss
\begin{equation}
    \mathbb{E}\!\left[ \mathit{l}_{\tau}\!\left( \mathbf{-1}\!\left( s \neq g \right) - \gamma || \phi(s') - \phi(g) || + || \phi(s) - \phi(g) ||\right) \right],
\end{equation}
where $\mathit{l}_{\tau}$ denotes the expectile loss \citep{newey1987expectile}, an asymmetric loss function that approximates the $\max$ operator in the Bellman backup~\citep{kostrikov2022iql}. This objective naturally supports the use of target networks~\citep{mnih2015dqn} and double estimators~\citep{van2016doubledqn} to improve learning stability. We included both in our implementation, following the original setup used by \citet{park2024foundation}.

The full set of hyperparameter values used to train the Hilbert models can be found in Table~\ref{tab:hilbert_params}.

\begin{table}[ht]
\caption{Hyperparameters used to train the Hilbert representation models.}
\label{tab:hilbert_params}
\centering
\begin{tabular}{ll}
\hline
\textbf{Hyperparameter}                      & \textbf{Value}                              \\ \hline
Latent Dimension                    & 32                            \\
Expectile                           & 0.95                           \\
Discount Factor                     & 0.99                          \\
Learning Rate                       & 0.0003                        \\
Target Network Smoothing Factor     & 0.005                         \\
Multi-Layer Perceptron Dimensions                      & (512, 512, 512) Fully-Connected Layers  \\
Activation Function (Hidden Layers) & GELU~\citep{hendrycks2016gelu}      \\
Layer Normalization (Hidden Layers) & True                                \\
Activation Function (Final Layer)   & Identity                            \\
Layer Normalization (Final Layer)   & False                               \\
Optimizer                           & Adam~\citep{kingma2015adam}         \\
Batch Size                          & 1024                               \\ \hline
\end{tabular}
\end{table}

\section{Ablation on choices of Delta and Kappa}
\label{app:abl-delta-kappa}

The sufficient threshold condition in Equation~\ref{eq:constants} is conservative. We therefore treat $\kappa$ and $\delta$ as an empirical hyperparameter in the experiments.

\begin{figure}[p]
    \centering
    \includegraphics[width=\linewidth]{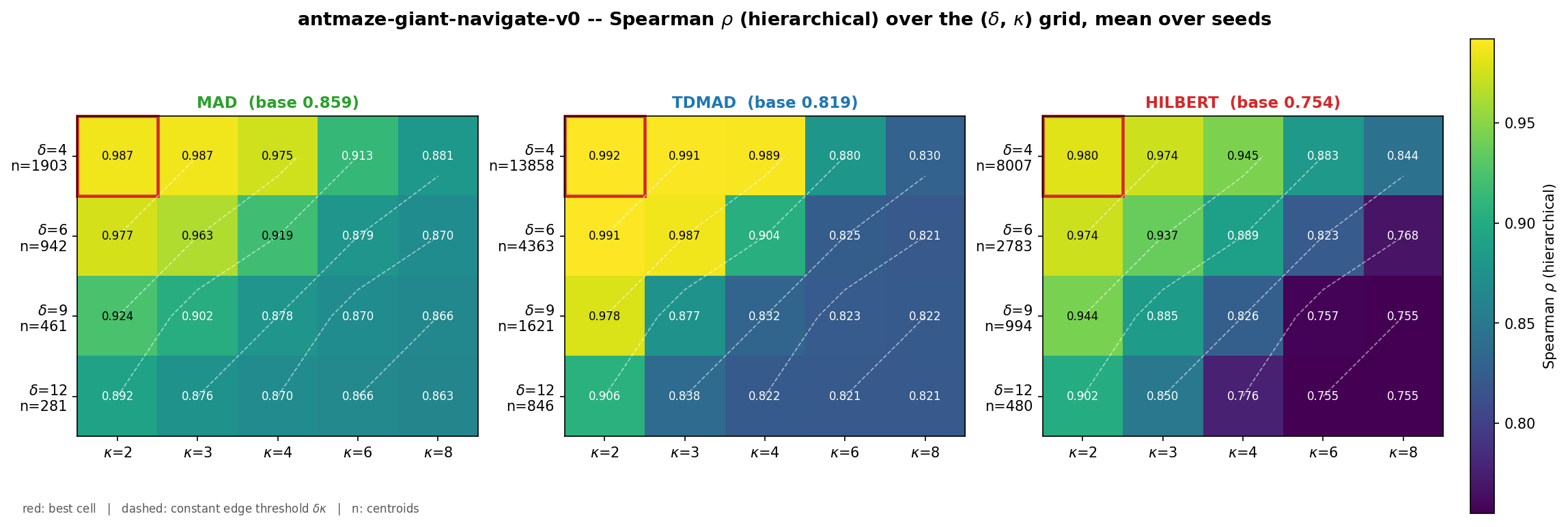}
    \\[6pt]
    \includegraphics[width=\linewidth]{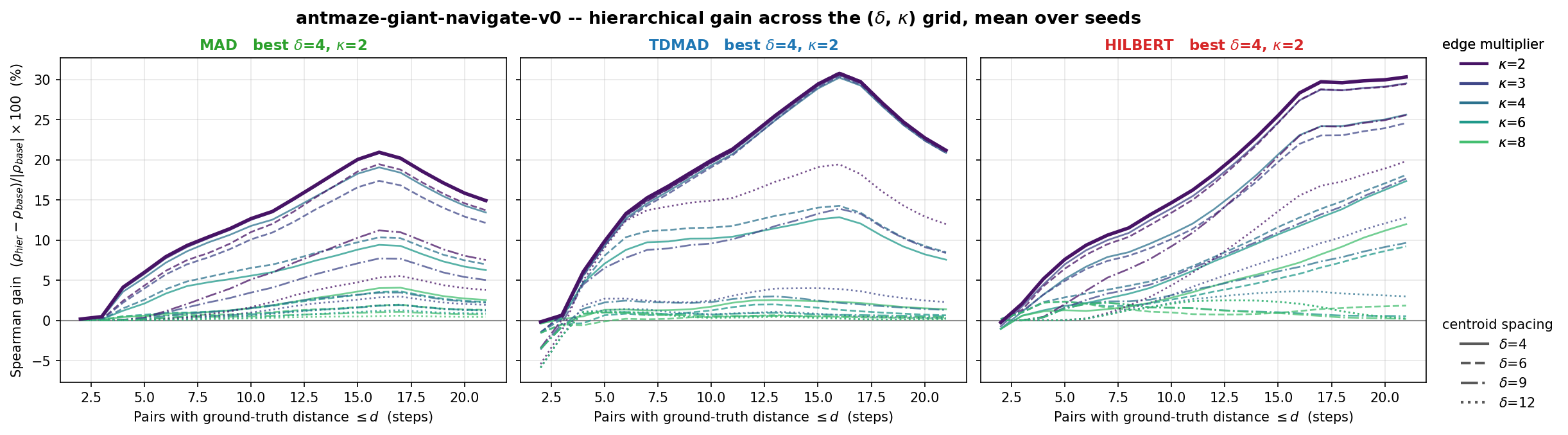}
    \caption{Ablation over the $(\delta,\kappa)$ grid on AntMaze Giant, \emph{navigate} split, for the three base models, averaged over seeds. \emph{Top:} Spearman correlation of the hierarchical estimate with the grid-maze distance; the base model's correlation is given in each panel title, the best cell is outlined in red, dashed lines are contours of constant edge threshold $\kappa\delta$, and $n$ is the number of landmarks selected at that spacing. \emph{Bottom:} relative gain over the base model as the evaluated range widens, with $\kappa$ shown by colour and $\delta$ by line style.}
    \label{fig:abl-am-giant}
\end{figure}

\begin{figure}[p]
    \centering
    \includegraphics[width=\linewidth]{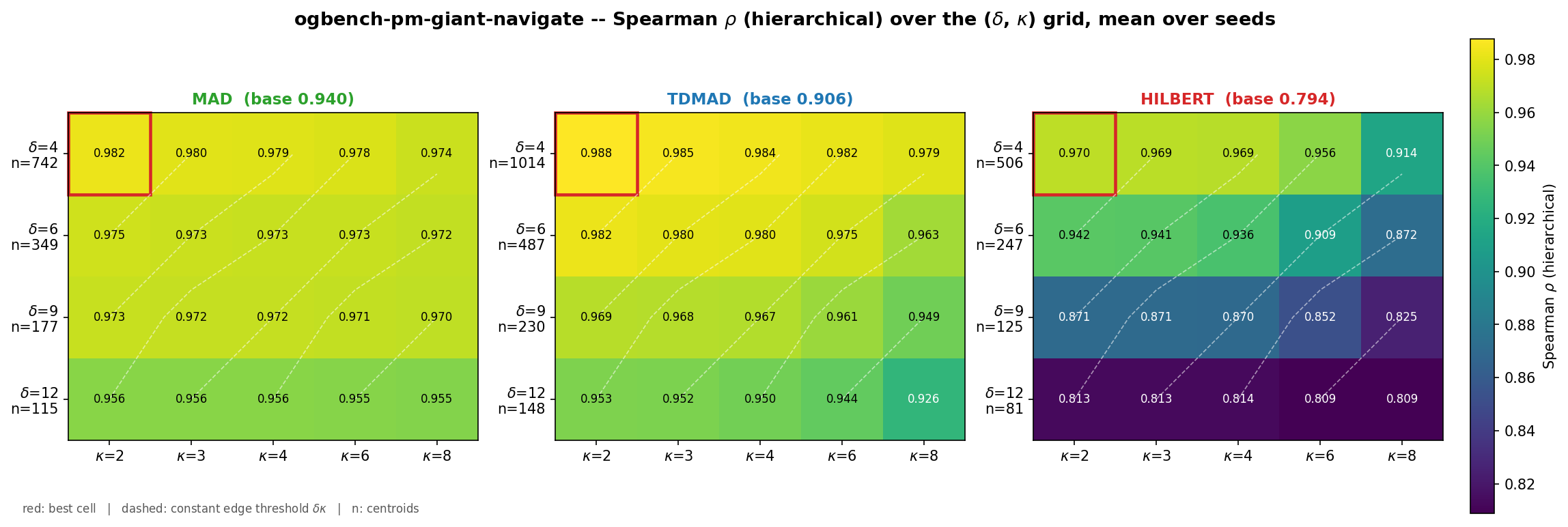}
    \\[6pt]
    \includegraphics[width=\linewidth]{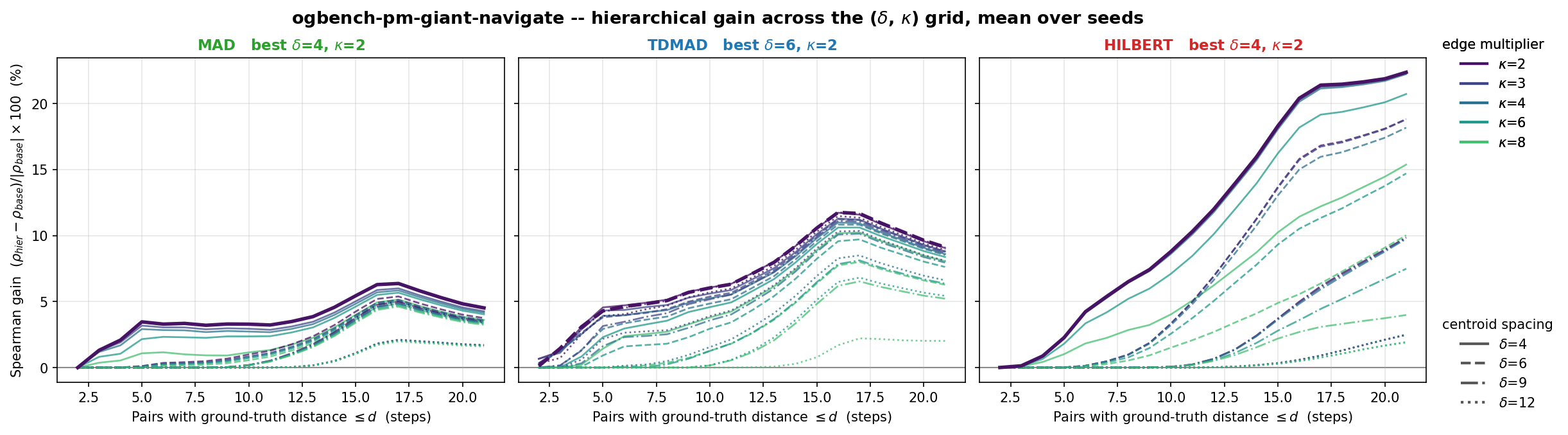}
    \caption{The same ablation on PointMaze Giant, \emph{navigate} split. Axes, colours, and annotations are as in Figure~\ref{fig:abl-am-giant}.}
    \label{fig:abl-pm-giant}
\end{figure}

\begin{figure}[p]
    \centering
    \includegraphics[width=\linewidth]{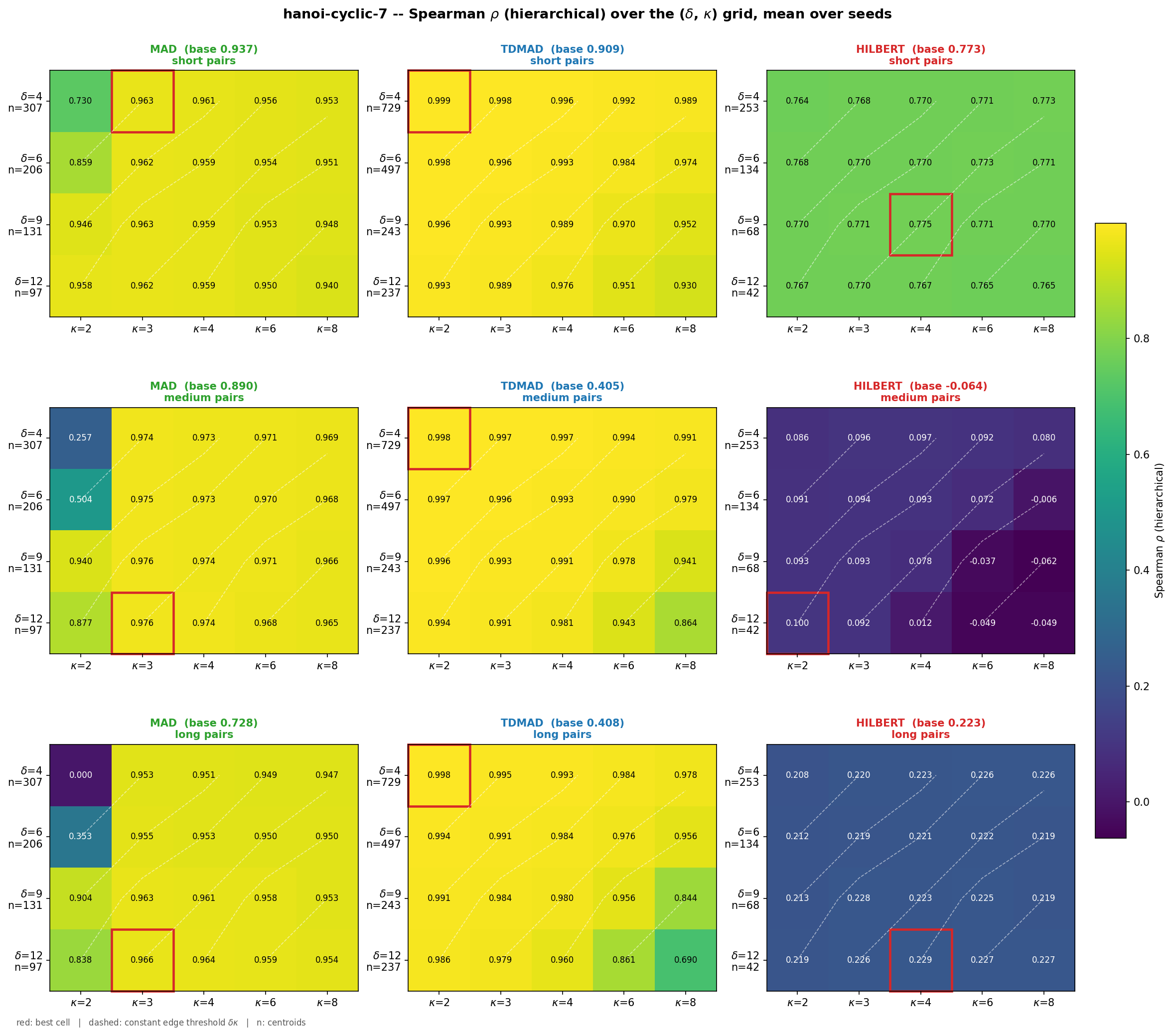}
    \caption{Ablation over the $(\delta,\kappa)$ grid on cyclic Towers of Hanoi, averaged over five seeds. Each panel shows the Spearman correlation of the hierarchical estimate with the exact MAD within one range (rows: short, medium, long pairs) for one base model (columns), with the base model's correlation in the panel title. The best cell is outlined in red, dashed lines are contours of constant edge threshold $\kappa\delta$, and $n$ is the number of landmarks. A cell with no routed pair in a range is shown as $0$.}
    \label{fig:hanoi-grid}
\end{figure}

We sweep the spacing $\delta\in\{4,6,9,12\}$ against the threshold multiplier $\kappa\in\{2,3,4,6,8\}$ on the \emph{giant} navigate datasets and on cyclic Towers of Hanoi, for all three base models and averaged over seeds. Figures~\ref{fig:abl-am-giant} and~\ref{fig:abl-pm-giant} report the resulting Spearman correlation over the whole $(\delta,\kappa)$ grid, together with the gain over the base model as the evaluated range widens. Figure~\ref{fig:hanoi-grid} reports the same grid for Towers of Hanoi, scored against the exact MAD for short, medium and long pairs.

On the mazes, the best cell is always $\delta=4$ with $\kappa=2$, and the correlation decreases as either parameter grows. As expected the effect of $\kappa$ is the stronger of the two, and it is most pronounced for the weakest base model: on AntMaze Giant with HILBERT the correlation falls from $0.980$ at $\kappa=2$ to $0.844$ at $\kappa=8$ with $\delta$ fixed at $4$, whereas for MAD on PointMaze Giant the same sweep moves it only from $0.982$ to $0.974$. This is the behaviour the analysis predicts qualitatively: $\kappa\delta$ is the radius over which the base model has to be accurate, so raising it admits more edges the base model can missclassify. Smaller $\delta$ also means more landmarks, shown as $n$ beside each row, so the preferred setting is the one that keeps every evaluation short at the cost of a larger graph.

In our experiment we use $\delta=4$ and $\kappa=2$ in all maze experiments, with one exception: the hierarchical TDMAD variant uses $\delta=6$. At $\delta=4$ that model's net grows to $13{,}858$ landmarks on AntMaze Giant, and becomes impractical. Raising the spacing to $\delta=6$ brings the net down to $4{,}363$ landmarks, about a thirtyfold saving in preprocessing, and costs almost nothing in quality.

\section{Planning algorithm}
\label{app:Planning}

Table~\ref{tab:horizon-vs-path} compares the planner horizon $H$ with the average number of decision stages required to reach the goal in each environment. In both AntMaze and PointMaze, $H$ covers only a fraction of the full goal-reaching horizon, so the planner must rely on the learned distance to successfully navigate towards distant goals.

\begin{table}[!ht]
\centering
\caption{Planning horizon compared to average steps to goal. The
goal is never inside the planner's lookahead, so the direction of travel comes
from the learned distance rather than from the simulated rollouts.}
\label{tab:horizon-vs-path}
\begin{tabular}{llrrr}
\toprule
Domain & Maze & $H$ (steps) & Steps to goal & $H$ / steps to goal \\
\midrule
\multirow{3}{*}{AntMaze}
  & Medium & 30 & 271 & 11.1\% \\
  & Large  & 30 & 451 & 6.7\% \\
  & Giant  & 30 & 863 & 3.5\% \\
\midrule
\multirow{3}{*}{PointMaze}
  & Medium & 5 & 184 & 2.7\% \\
  & Large  & 5 & 316 & 1.6\% \\
  & Giant  & 5 & 542 & 0.9\% \\
\bottomrule
\end{tabular}
\end{table}

\paragraph{Planning in PointMaze.}
We validate distance quality using simple random shooting. At each replanning step, we sample $50$ actions uniformly from a $600$-bin grid ($24$ per dimension), simulate each for $H=5$ steps, and execute the action whose trajectory minimizes learned distance to the goal. No refinement, elite selection, or smoothing is applied. This deliberately weak planner ensures planning success depends on distance accuracy.

\paragraph{Planning in AntMaze.}
Naive random sampling fails in AntMaze's 8D torque space, sampling 240 free parameters yields uncoordinated walking. Instead, we sample a central pattern generator (CPG) with 25 parameters: amplitude, phase, and offset per joint, plus a shared frequency. These parameters decode into a 30-step torque sequence:
\begin{equation}
a_j(t) \;=\; \mathrm{offset}_j + \mathrm{amp}_j \sin(2\pi f t + \mathrm{phase}_j) .
\end{equation}
This reduces the search space by an order of magnitude while restricting sampling to locomotion-compatible sequences. We refine with 3 CEM iterations (keeping 10 best candidates), warm-start the gait across replans.

\paragraph{Comparability.}
Both domains use identical planners and budgets when comparing flat vs.\ hierarchical distances. Any success rate difference is thus due to distance quality, not planner differences. The extra machinery in AntMaze (CPG, CEM refinement) is necessary for the dynamics, not an advantage for any particular method.

\section{Additional Planning Results}
\label{app:planning-results}

This appendix reports the planning success rate of Table~\ref{tab:planning} for every environment, base distance model, and dataset split.

\begin{table}[p]
  \centering
  \caption{Planning success rate on every environment, over 50 episodes, reported as mean $\pm$ standard deviation over five seeds. Columns are as in Table~\ref{tab:planning}, whose \emph{giant} rows are repeated here so that the comparison across maze sizes can be read in one place.}
  \label{tab:planning-full}
  \small
  \setlength{\tabcolsep}{4pt}
  \renewcommand{\arraystretch}{0.95}
  \begin{tabular}{llrrr}
    \toprule
    Environment & Base / Data & Flat & Hierarchical & $\Delta$ \\
    \midrule
    PointMaze Medium & HILBERT / navigate & $0.408 \pm 0.152$ & $1.000 \pm 0.000$ & $+0.592$ \\
     & HILBERT / stitch & $0.360 \pm 0.113$ & $1.000 \pm 0.000$ & $+0.640$ \\
     & MAD / navigate & $0.552 \pm 0.041$ & $1.000 \pm 0.000$ & $+0.448$ \\
     & MAD / stitch & $0.372 \pm 0.094$ & $1.000 \pm 0.000$ & $+0.628$ \\
     & TDMAD / navigate & $0.388 \pm 0.098$ & $1.000 \pm 0.000$ & $+0.612$ \\
     & TDMAD / stitch & $0.380 \pm 0.174$ & $1.000 \pm 0.000$ & $+0.620$ \\
    \midrule
    PointMaze Large & HILBERT / navigate & $0.056 \pm 0.050$ & $1.000 \pm 0.000$ & $+0.944$ \\
     & HILBERT / stitch & $0.040 \pm 0.070$ & $1.000 \pm 0.000$ & $+0.960$ \\
     & MAD / navigate & $0.324 \pm 0.145$ & $1.000 \pm 0.000$ & $+0.676$ \\
     & MAD / stitch & $0.476 \pm 0.084$ & $1.000 \pm 0.000$ & $+0.524$ \\
     & TDMAD / navigate & $0.308 \pm 0.082$ & $1.000 \pm 0.000$ & $+0.692$ \\
     & TDMAD / stitch & $0.272 \pm 0.081$ & $1.000 \pm 0.000$ & $+0.728$ \\
    \midrule
    PointMaze Giant & HILBERT / navigate & $0.000 \pm 0.000$ & $0.992 \pm 0.016$ & $+0.992$ \\
     & HILBERT / stitch & $0.000 \pm 0.000$ & $0.984 \pm 0.023$ & $+0.984$ \\
     & MAD / navigate & $0.072 \pm 0.068$ & $0.984 \pm 0.023$ & $+0.912$ \\
     & MAD / stitch & $0.040 \pm 0.062$ & $1.000 \pm 0.000$ & $+0.960$ \\
     & TDMAD / navigate & $0.028 \pm 0.039$ & $1.000 \pm 0.000$ & $+0.972$ \\
     & TDMAD / stitch & $0.000 \pm 0.000$ & $1.000 \pm 0.000$ & $+1.000$ \\
    \midrule
    AntMaze Medium & HILBERT / explore & $0.104 \pm 0.079$ & $0.992 \pm 0.010$ & $+0.888$ \\
     & HILBERT / navigate & $0.224 \pm 0.069$ & $0.956 \pm 0.050$ & $+0.732$ \\
     & HILBERT / stitch & $0.320 \pm 0.090$ & $0.840 \pm 0.062$ & $+0.520$ \\
     & MAD / explore & $0.492 \pm 0.041$ & $0.996 \pm 0.008$ & $+0.504$ \\
     & MAD / navigate & $0.484 \pm 0.053$ & $1.000 \pm 0.000$ & $+0.516$ \\
     & MAD / stitch & $0.428 \pm 0.107$ & $0.996 \pm 0.008$ & $+0.568$ \\
     & TDMAD / explore & $0.296 \pm 0.078$ & $1.000 \pm 0.000$ & $+0.704$ \\
     & TDMAD / navigate & $0.236 \pm 0.086$ & $0.996 \pm 0.008$ & $+0.760$ \\
     & TDMAD / stitch & $0.300 \pm 0.069$ & $1.000 \pm 0.000$ & $+0.700$ \\
    \midrule
    AntMaze Large & HILBERT / navigate & $0.036 \pm 0.043$ & $0.536 \pm 0.148$ & $+0.500$ \\
     & HILBERT / stitch & $0.024 \pm 0.023$ & $0.488 \pm 0.048$ & $+0.464$ \\
     & MAD / navigate & $0.340 \pm 0.063$ & $0.972 \pm 0.020$ & $+0.632$ \\
     & MAD / stitch & $0.308 \pm 0.045$ & $0.980 \pm 0.018$ & $+0.672$ \\
     & TDMAD / navigate & $0.328 \pm 0.041$ & $0.984 \pm 0.008$ & $+0.656$ \\
     & TDMAD / stitch & $0.324 \pm 0.008$ & $0.988 \pm 0.016$ & $+0.664$ \\
    \midrule
    AntMaze Giant & HILBERT / navigate & $0.000 \pm 0.000$ & $0.540 \pm 0.092$ & $+0.540$ \\
     & HILBERT / stitch & $0.000 \pm 0.000$ & $0.388 \pm 0.057$ & $+0.388$ \\
     & MAD / navigate & $0.004 \pm 0.008$ & $0.992 \pm 0.016$ & $+0.988$ \\
     & MAD / stitch & $0.016 \pm 0.015$ & $0.964 \pm 0.023$ & $+0.948$ \\
     & TDMAD / navigate & $0.000 \pm 0.000$ & $0.980 \pm 0.013$ & $+0.980$ \\
     & TDMAD / stitch & $0.000 \pm 0.000$ & $0.984 \pm 0.015$ & $+0.984$ \\
    \bottomrule
  \end{tabular}
\end{table}

\section{Additional Distance-Quality Results}
\label{app:distance-results}

This appendix reports the Spearman-correlation comparison of Figure~\ref{fig:sp_giant_stitch} for every remaining environment and dataset split.

\begin{figure}[p]
    \centering
    \includegraphics[width=\linewidth]{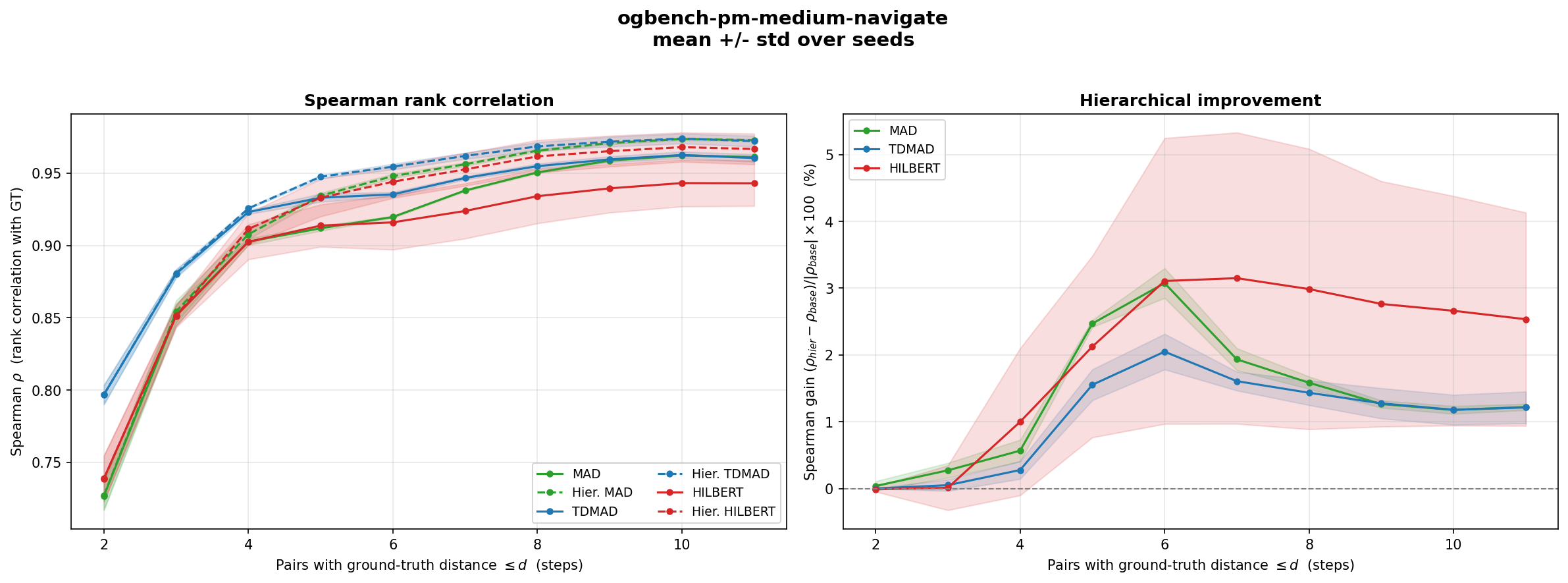}
    \\[4pt]
    \includegraphics[width=\linewidth]{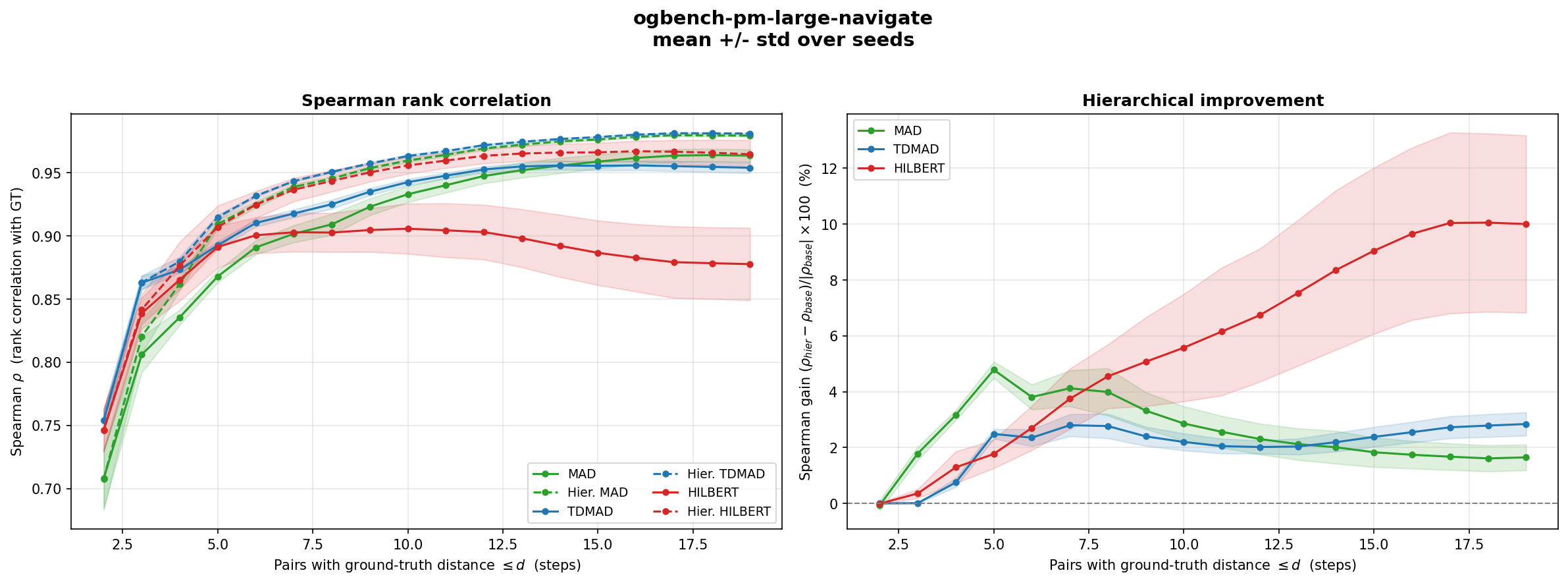}
    \\[4pt]
    \includegraphics[width=\linewidth]{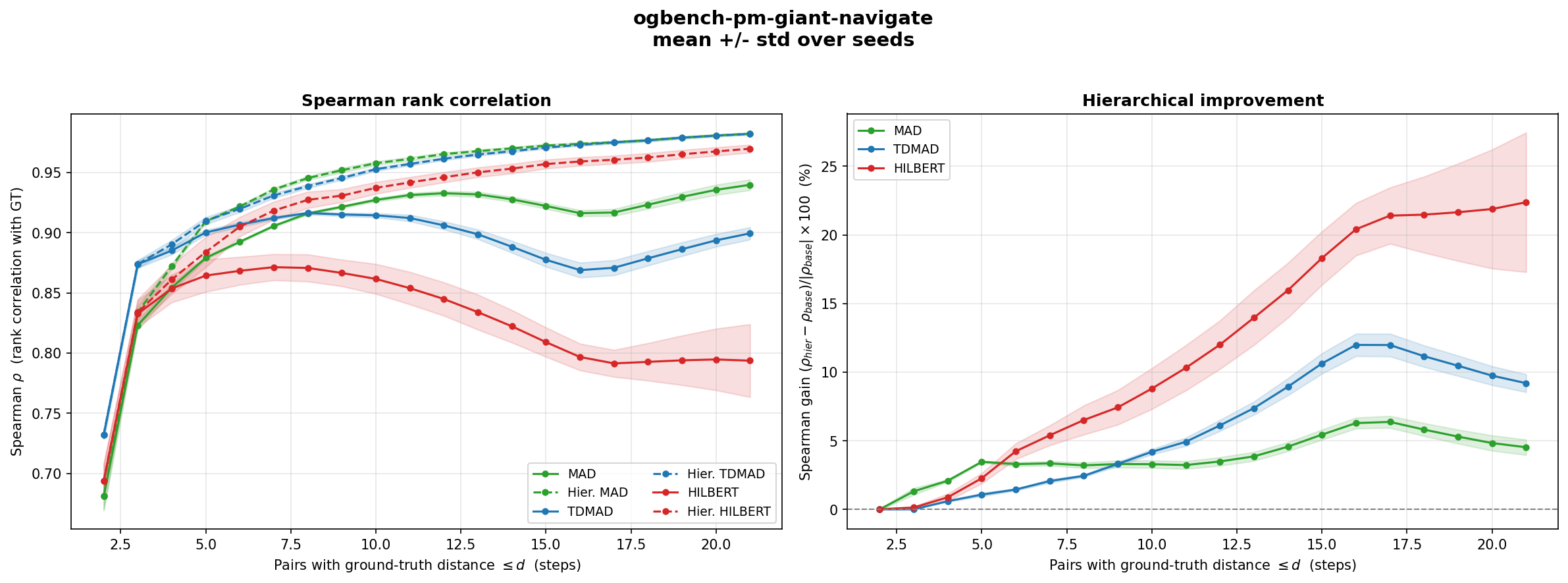}
    \caption{Spearman correlation with grid-geodesic distance as the evaluated range increases (left) and relative improvement from the hierarchy (right). Results are mean $\pm$ standard deviation over five seeds for all three base models. PointMaze, \emph{navigate} split, at medium, large, and giant sizes.}
    \label{fig:sp_pm_navigate}
\end{figure}

\begin{figure}[p]
    \centering
    \includegraphics[width=\linewidth]{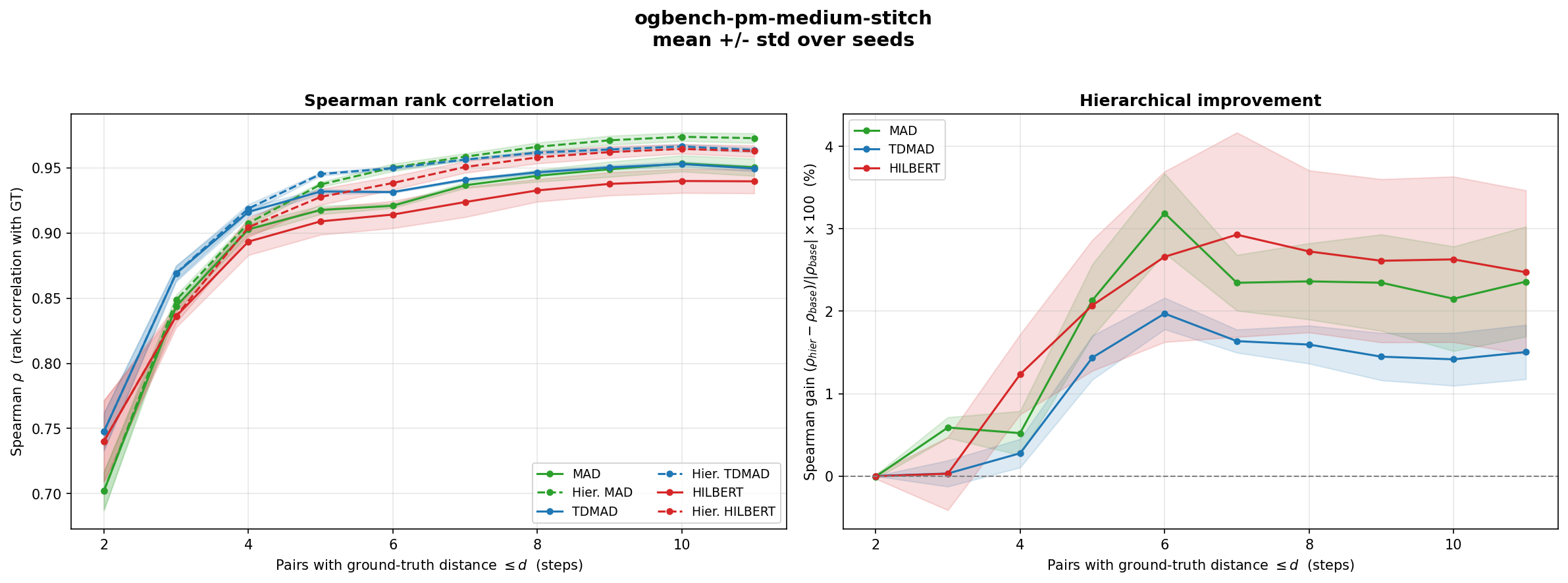}
    \\[4pt]
    \includegraphics[width=\linewidth]{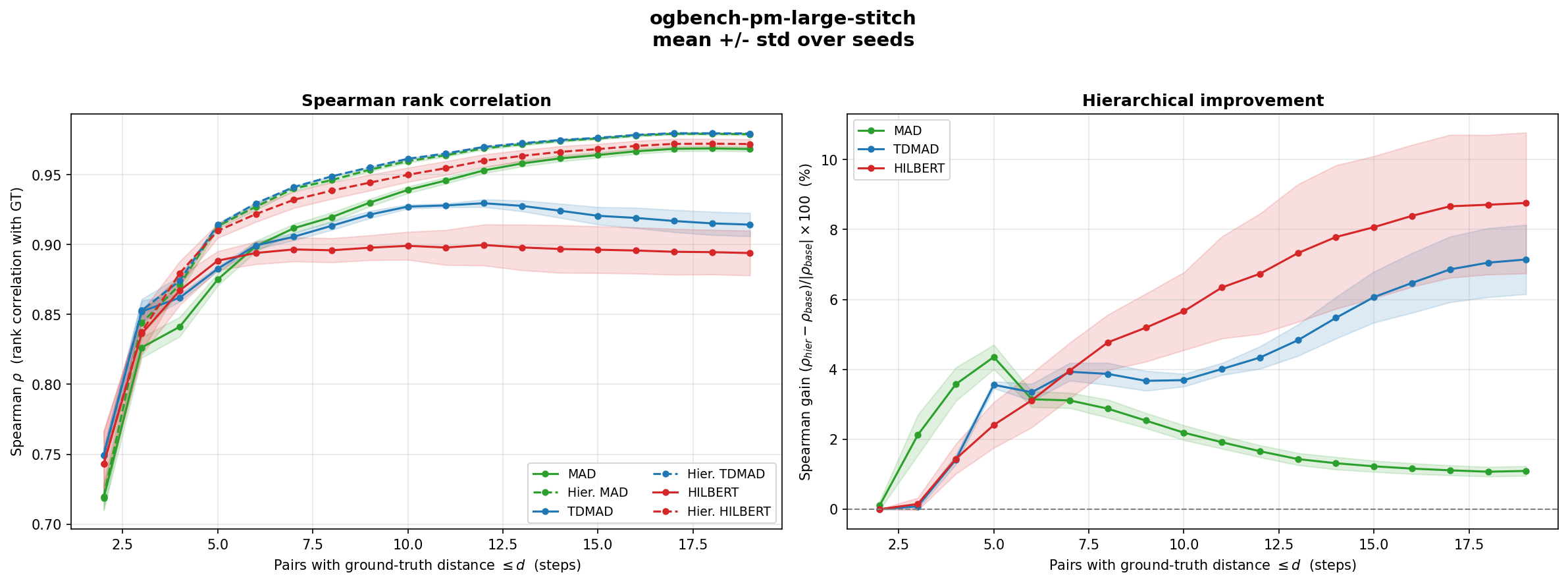}
    \\[4pt]
    \includegraphics[width=\linewidth]{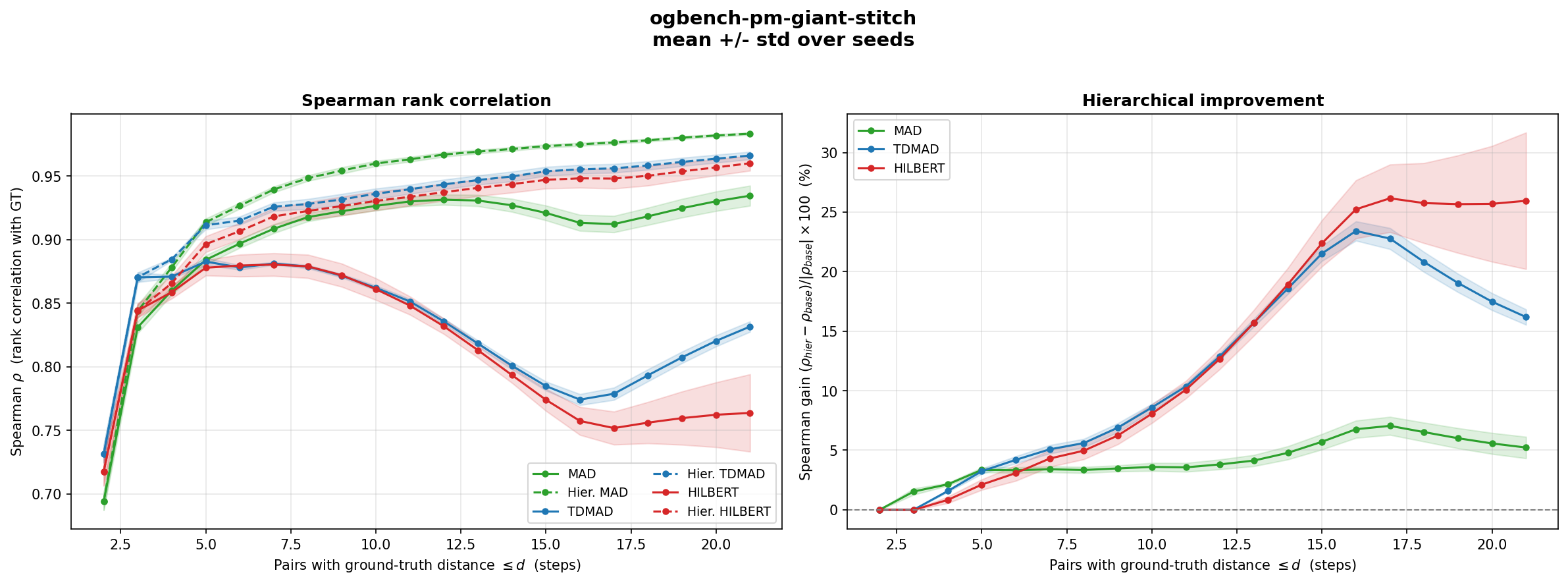}
    \caption{Spearman correlation with grid-geodesic distance as the evaluated range increases (left) and relative improvement from the hierarchy (right). Results are mean $\pm$ standard deviation over five seeds for all three base models. PointMaze, \emph{stitch} split.}
    \label{fig:sp_pm_stitch}
\end{figure}

\begin{figure}[p]
    \centering
    \includegraphics[width=\linewidth]{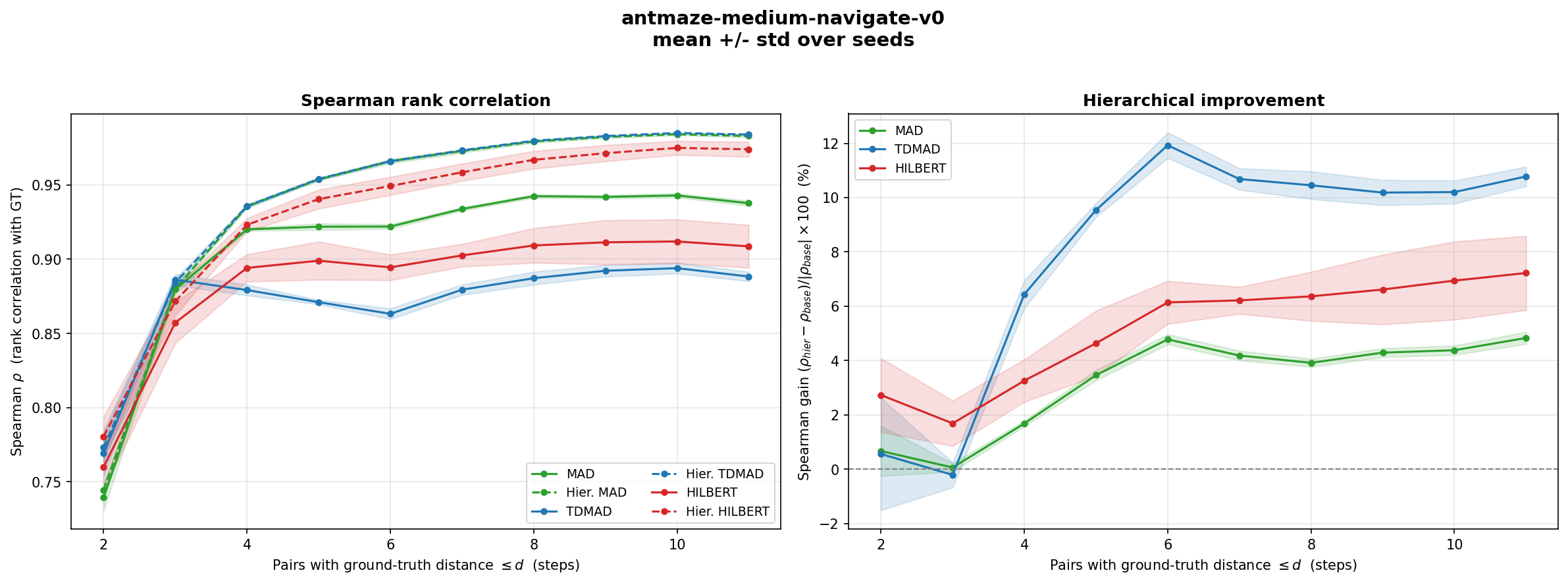}
    \\[4pt]
    \includegraphics[width=\linewidth]{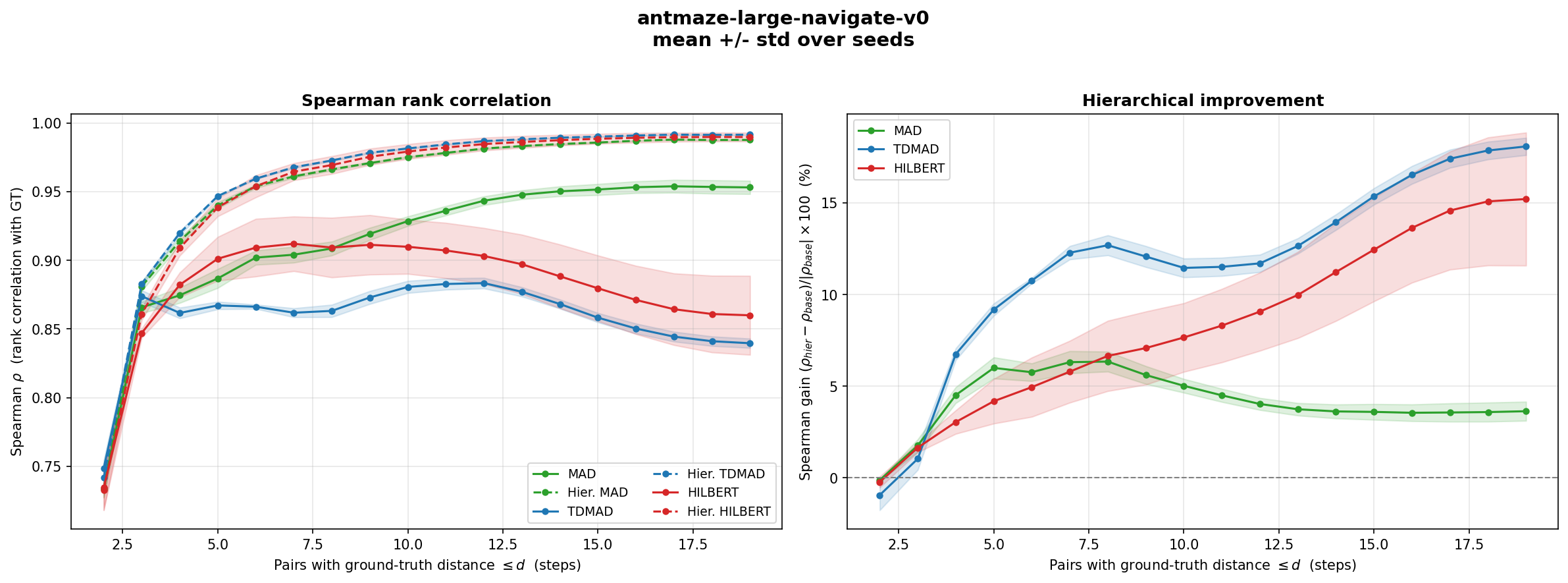}
    \\[4pt]
    \includegraphics[width=\linewidth]{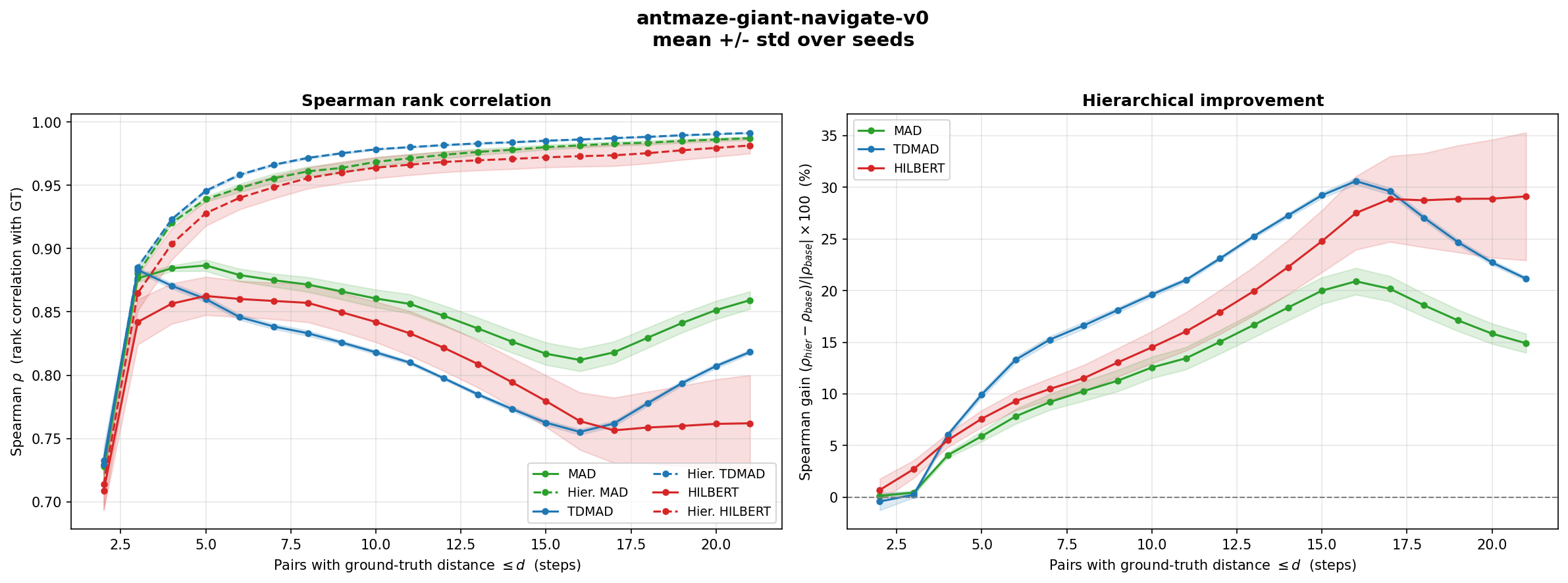}
    \caption{Spearman correlation with grid-geodesic distance as the evaluated range increases (left) and relative improvement from the hierarchy (right). Results are mean $\pm$ standard deviation over five seeds for all three base models. AntMaze, \emph{navigate} split.}
    \label{fig:sp_am_navigate}
\end{figure}

\begin{figure}[p]
    \centering
    \includegraphics[width=\linewidth]{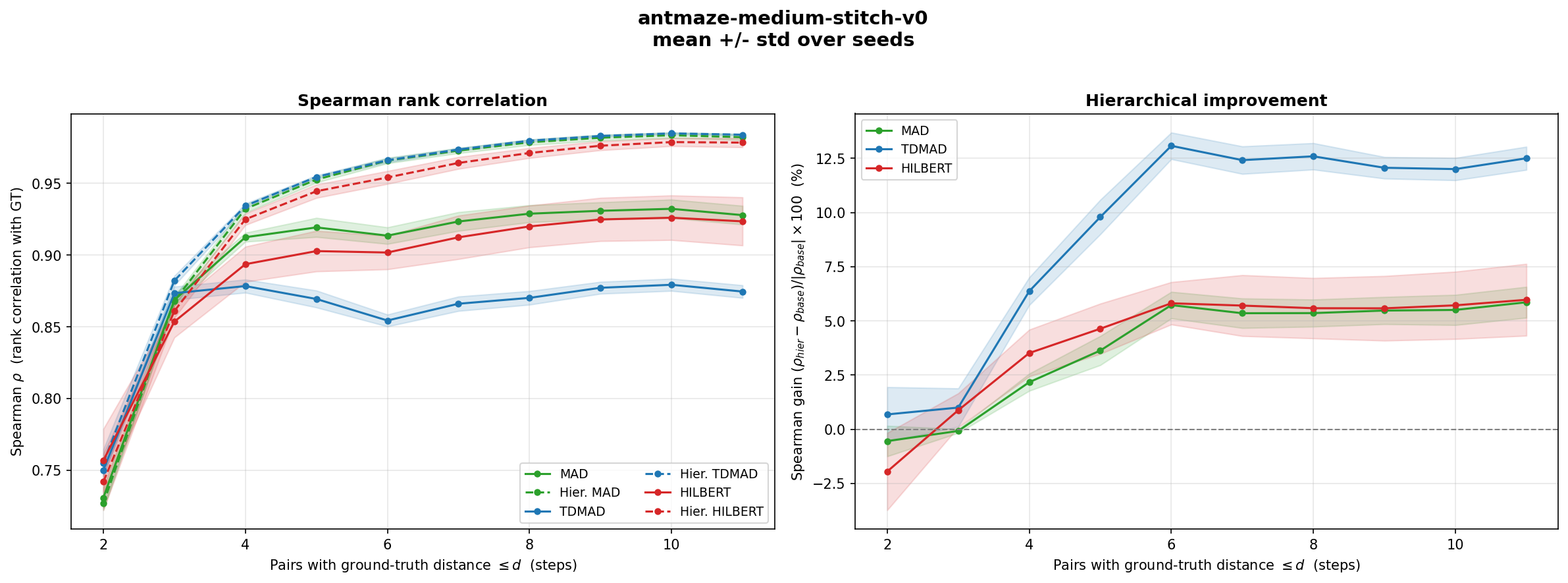}
    \\[4pt]
    \includegraphics[width=\linewidth]{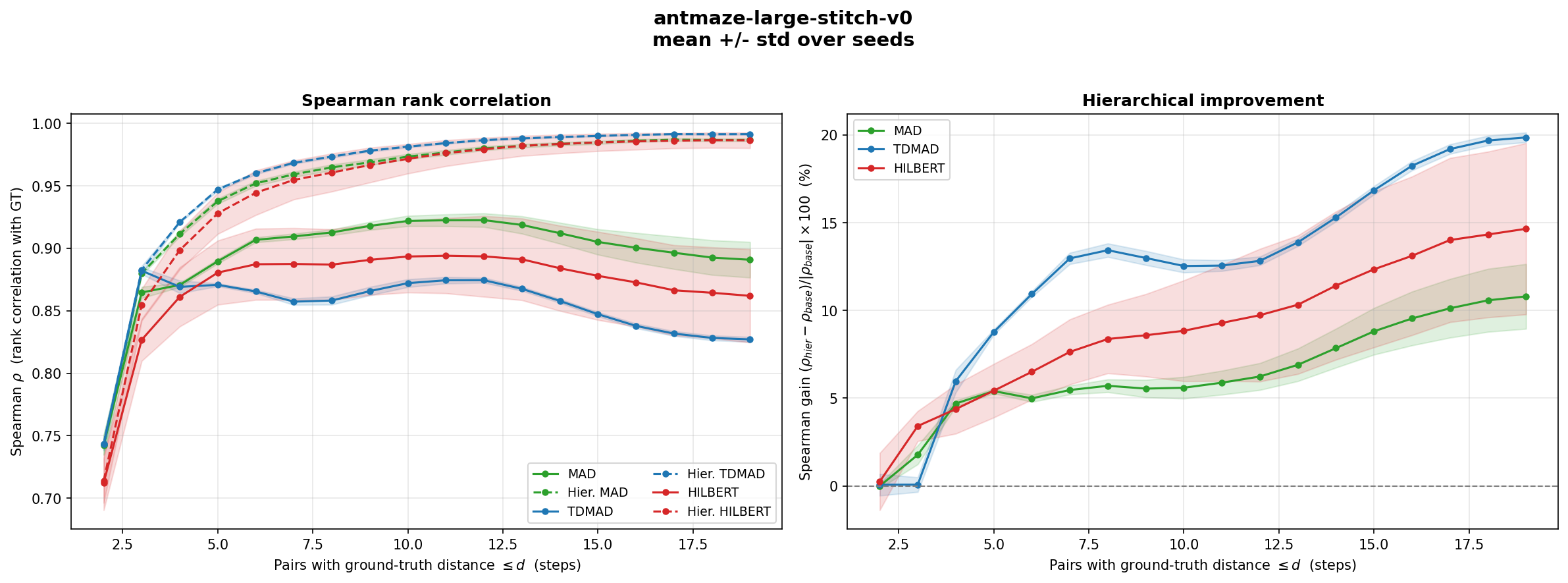}
    \\[4pt]
    \includegraphics[width=\linewidth]{images/am-giant-stitch.png}
    \caption{Spearman correlation with grid-geodesic distance as the evaluated range increases (left) and relative improvement from the hierarchy (right). Results are mean $\pm$ standard deviation over five seeds for all three base models. AntMaze, \emph{stitch} split.}
    \label{fig:sp_am_stitch}
\end{figure}

\begin{figure}[p]
    \centering
    \includegraphics[width=\linewidth]{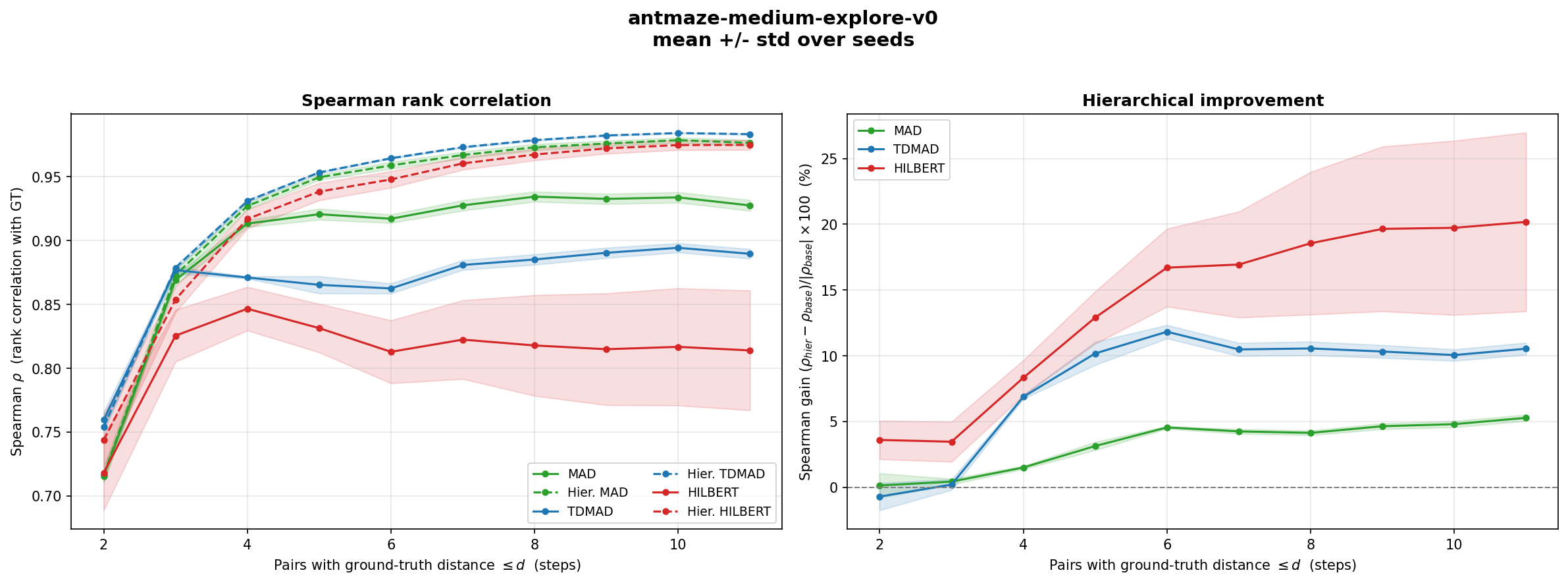}
    \caption{Spearman correlation with grid-geodesic distance as the evaluated range increases (left) and relative improvement from the hierarchy (right). Results are mean $\pm$ standard deviation over five seeds for all three base models. AntMaze Medium, \emph{explore} split.}
    \label{fig:sp_am_explore}
\end{figure}

\section{Cyclic Towers of Hanoi}
\label{app:hanoi}

The mazes of Section~\ref{sec:results} have no exact MAD, so distance quality there is measured against the grid-geodesic distance. This appendix adds a discrete domain where the MAD is known exactly for every ordered pair of states.

\paragraph{Environment.}
In the cyclic Towers of Hanoi~\citep{atkinson1981cyclic}, disks of different sizes sit on three pegs. A move takes the top disk of one peg to another in a clockwise order, and a disk may never rest on a smaller one. We use seven disks. A state assigns each disk to a peg, which gives $3^7=2{,}187$ states. There are three actions, one per clockwise move, and an illegal move leaves the state unchanged. The transition graph is directed and strongly connected. We compute the exact MAD $d^*$ for every ordered pair by breadth-first search; its diameter is $1{,}223$ actions.

\paragraph{Asymmetry.}
The MAD here is strongly asymmetric. Undoing a single move costs between $2$ and $983$ actions, depending on the disk that moved. A symmetric base model cannot satisfy Assumption~\ref{ass:local} on this domain. 

\paragraph{Data and base models.}
Each seed collects $5{,}000$ trajectories of $100$ uniformly random actions, each from a uniformly random start state. We train MAD, TDMAD, and HILBERT on these trajectories with the hyperparameters of Appendix~\ref{app:base-models}, over five seeds, and build each hierarchy from the same trajectories.

\paragraph{Hierarchy.}
We keep the spacings of Section~\ref{sec:results}, $\delta=4$ for MAD and HILBERT and $\delta=6$ for TDMAD, together with the max selection rule. The one change is the threshold multiplier, which is $\kappa=3$ instead of $\kappa=2$.

\paragraph{Metrics.}
We score all $4{,}780{,}782$ ordered pairs of distinct states and split them by $d^*(x,y)$ into thirds of the diameter: short ($d^*\le408$), medium ($408<d^*\le815$), and long ($d^*>815$). Within each range we report the Spearman correlation with $d^*$. We also report \emph{asymmetry accuracy}. For an estimate $\hat d$, either the base model $d_\theta$ or the hierarchical estimate $\hat d_H$, it is the fraction of ordered pairs with $d^*(x,y)\neq d^*(y,x)$ for which $\hat d(x,y)-\hat d(y,x)$ has the same sign as $d^*(x,y)-d^*(y,x)$, with ties counted as one half. Any symmetric model therefore scores exactly $0.5$.

\begin{figure}[t]
    \centering
    \includegraphics[width=\linewidth]{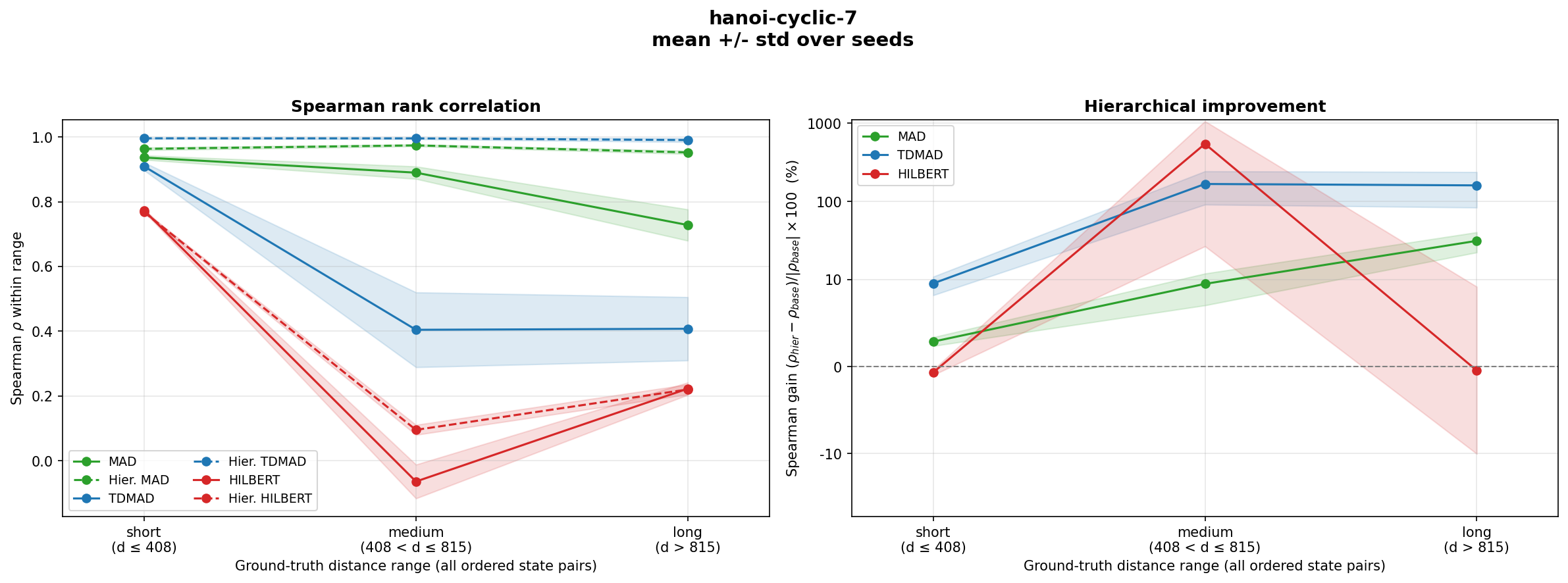}
    \caption{Cyclic Towers of Hanoi with seven disks, scored against the exact MAD over all ordered state pairs. \emph{Left:} Spearman correlation within each third of the diameter for the base models (solid) and their hierarchies (dashed). \emph{Right:} relative gain from the hierarchy; HILBERT's medium-range gain is large because its base correlation is close to zero. Mean $\pm$ standard deviation over five seeds.}
    \label{fig:hanoi-spearman}
\end{figure}

\paragraph{Results.}
Figure~\ref{fig:hanoi-spearman} shows the pattern seen on the mazes. The base models lose rank agreement as the range grows: from short to long pairs, MAD falls from $0.937$ to $0.728$ and TDMAD from $0.909$ to $0.408$. The hierarchical estimates stay at or above $0.965$ in every range for both models, so the relative gain is largest on medium and long pairs.

\end{document}